%% file: Transfer_Learning.tex
\RequirePackage{amsmath}
\documentclass[runningheads]{llncs}
\usepackage{graphicx}
\usepackage{subcaption}
\usepackage{mathtools}
\usepackage{hyperref} 
\usepackage{multirow,multicol}
\usepackage{amssymb}
\usepackage{pgf}

\begin{document}
\mainmatter              

\title{ Transfer of Pretrained Model Weights Substantially Improves Semi-Supervised Image Classification}
\titlerunning{Transfer of Pretrained Model  Improves Semi-Supervised Learning }  

%
\author{Attaullah Sahito,  Eibe Frank, \and Bernhard Pfahringer 
}
\authorrunning{Attaullah et al.} 
%
\tocauthor{Attaullah Sahito, Eibe Frank, and Bernhard Pfahringer
}

\institute{Department of Computer Science, University of Waikato, Hamilton, New Zealand.\\
\email{a19@students.waikato.ac.nz}, \email{\{eibe,bernhard\}@waikato.ac.nz}}

\maketitle

\begin{abstract}

Deep neural networks produce state-of-the-art results when trained on a large number of labeled examples but tend to overfit when small amounts of labeled examples are used for training. Creating a large number of labeled examples requires considerable resources, time, and effort. If labeling new data is not feasible, so-called semi-supervised learning can achieve better generalisation than purely supervised learning by employing unlabeled instances as well as labeled ones. The work presented in this paper is motivated by the observation that transfer learning provides the opportunity to potentially further improve performance by exploiting models pretrained on a similar domain. More specifically, we explore the use of transfer learning when performing semi-supervised learning using self-learning. The main contribution is an empirical evaluation of transfer learning using different combinations of similarity metric learning methods and label propagation algorithms in semi-supervised learning. We find that transfer learning always substantially improves the model's accuracy when few labeled examples are available, regardless of the type of loss used for training the neural network. This finding is obtained by performing extensive experiments on the SVHN, CIFAR10, and Plant Village image classification datasets and applying pretrained weights from Imagenet for transfer learning.

\end{abstract}

\section{Introduction}
Neural networks are frequently used for image classification tasks and yield state-of-the-art results in this application. However, for training, these models generally need a lot of labeled samples, and they tend to overfit on small amounts of labeled data. This problem is of particular importance when limited labeled samples are available due to time or financial constraints. Addressing this problem requires machine learning methods that are able to work with a limited amount of labeled data and also make efficient use of the side information available from unlabeled data.

Semi-supervised learning (SSL) aims to improve performance by exploiting both  labeled and unlabeled examples. Given an input space $X$ containing the examples, SSL methods are designed to  work with labeled examples $L=\{(x_1,y_1),(x_2,y_2),...,(x_{|L|},y_{|L|})\}$ and unlabeled examples $U= \{{x^{'}_1},{x^{'}_2},...,{x^{'}_{|U|}} \}$, where $x_i, x^{'}_{j} \in X$ with $i= 1,2,...,|L|$  and $j= 1,2,...,|U|$  and $y_i$ are the labels of $x_i$, with $y_i \in \{1,2,3,...,c\}$  ($c$ being the number of classes). 

A few assumptions are required to make semi-supervised learning a principled approach~\cite{chapelle2006semi}:
\begin{enumerate}
    \item If two instances $x_1 , x_2$ are close in a high-density region, then their corresponding outputs $y_1 , y_2$ should also be close.
    \item If instances are in the same structure (referred to as a cluster or  manifold), they are likely to be of the same class.
    \item The decision boundary between classes should lie in a low-density region of the input space.
\end{enumerate}

Almost all standard neural networks for image classification are trained by minimising cross-entropy loss on  labeled training data. In this paper, along with cross-entropy loss, we also consider another class of losses, comprising so-called similarity metric learning losses, which operate on the relationships between samples such that instances of the same class are considered similar and those belonging to different classes are considered dissimilar. Once a similarity function has been trained, which is parameterised by a neural network, feature vectors (embeddings) of examples produced by the network will be grouped together according to class labels, normally in Euclidean space. These learned embeddings lend themselves naturally to semi-supervised learning because they can be employed to assign class labels to unlabeled examples using very simple classification methods such as nearest-neighbor classifiers.

\begin{figure}
	\centering
	\includegraphics[scale=0.50] {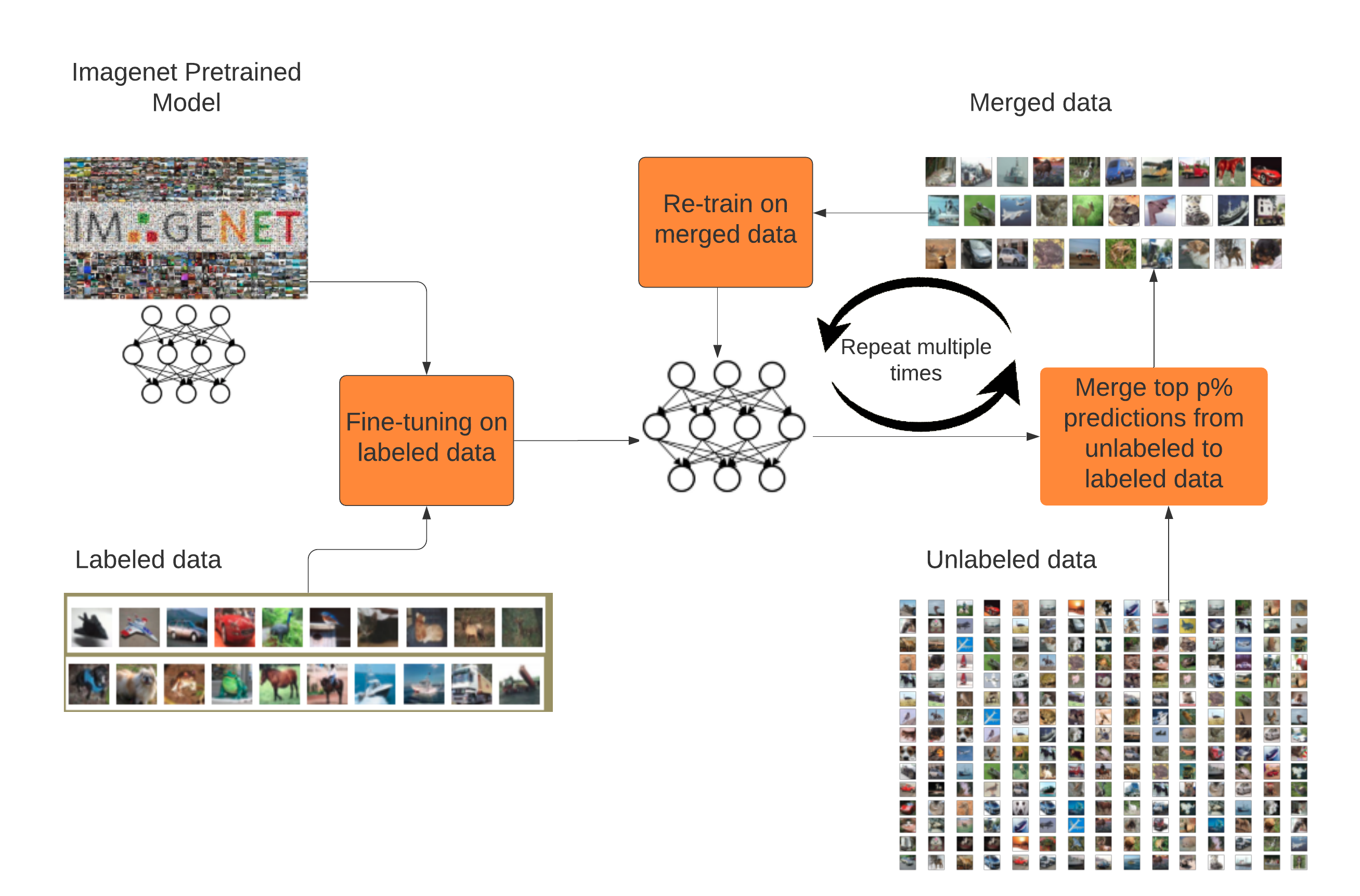}
	\caption{Overview of the approach.}
	\label{fig:transfer}
\end{figure}

This approach is related to work on pseudo-labeling~\cite{lee2013pseudo,rosenberg2007semi}, where the model is initially trained on limited data. However, in this paper, instead of applying random initialisation of network parameters when training starts, we investigate using pretrained weights from another domain and show that this provides much better generalisation ability. Using pretrained model weights is a standard approach for transfer learning in supervised settings, but appears to have received little attention in the context of semi-supervised learning, particularly when applying self-learning with metric learning. 

We use a pretrained neural network model trained on Imagenet~\cite{russakovsky2015imagenet}. A schematic overview of the proposed approach is shown in Figure~\ref{fig:transfer}. Fine-tuning on data from the target domain is performed on the (very small) initial set of labeled examples. Following that, confident predictions for unlabeled examples are added to labeled examples for iterative retraining of the neural network---this is the standard self-learning method for semi-supervised learning. It enables us to obtain more labeled training data and the assumption is that this eventually helps in achieving significant performance improvements. In our experiments on image classification tasks, we compare using pretrained weights for the neural network to random initialisation of the weights.

The main contribution of this work is an extensive empirical investigation of transfer learning in the context of self-learning. Using cross-entropy loss as well as combinations of similarity metric learning losses (e.g., triplet loss, contrastive loss, and Arcface loss) with simple nearest-neighbor-based label propagation, we find that transfer learning always substantially improves the classification accuracy of the model when few labeled examples are available, regardless of which loss function is used for training the neural network. More specifically, for semi-supervised learning using self-learning on the SVHN, CIFAR10, and Plant Village image classification datasets, we obtain a substantial improvement using pretrained weights when few labeled examples are available for training.  Thus, our results indicate that the well-established method of performing transfer learning by re-using pretrained weights---commonly applied when performing a purely supervised training of a neural network---is particularly useful in the context of semi-supervised learning.

\section{Related Work}
In this section, we briefly discuss some existing work on semi-supervised learning and transfer learning.
\subsection{Semi-supervised Learning}
Semi-supervised Learning (SSL) lies between supervised and unsupervised learning. SSL tries to  employ labeled examples as well as unlabeled examples for more accurate prediction.
There are many different techniques available from the literature  on SSL using deep neural networks. Some employ autoencoders~\cite{maaloe2016auxiliary,rasmus2015semi}, others use generative models~\cite{dai2017good,wei2018improving,salimans2016improved} or are based on regularization ideas~\cite{miyato2018virtual,sajjadi2016regularization}. In pseudo-labeling~\cite{lee2013pseudo}, the model is trained on the limited labeled data first and then re-trained on an extended set of labeled data, based on the predictions of the original model for the unlabelled training data.

Our method builds on work investigating transferring learning using both cross-entropy loss and similarity-based metric learning with neural networks. Pair and triplet based loss functions provide the foundation for standard approaches to metric learning. A classic pair-based method is to use contrastive loss~\cite{chopra2005learning}, which tries to bring similar pairs closer and push farther away dissimilar pairs. Pairs can be extended to triplets. They consist of an anchor, a positive, and a negative example, where the anchor is more similar to the positive example than the negative one. The resulting triplet loss function~\cite{schroff2015facenet} was originally used on triplets of images for face verification. Metric learning-based loss functions~\cite{weston2012deep} have also been successfully employed for image classification.

Another related class of metric learning methods are based on modified classification losses. Examples include Arcface~\cite{deng2019arcface}, Sphereface~\cite{liu2017sphereface} and Cosface~\cite{wang2018cosface}. For metric learning, Arcface, Sphereface, and Cosface  apply multiplicative-angular, additive-cosine, and additive-angular margins, respectively. 

\subsection{Transfer Learning}
Since the successful Imagenet challenge~\cite{russakovsky2015imagenet}, transfer learning has been used widely in visual recognition tasks such as object detection~\cite{girshick2014rich}. Transfer learning uses the network weights learned by training on the large and labeled Imagenet dataset and fine-tunes the weights for the respective target domain. When the target domain is sufficiently closely related to the source domain of Imagenet, then transfer learning usually generalizes much better than training from scratch on the smaller target domain alone.

\section{Semi-supervised Learning using Self-learning}

The semi-supervised learning approach we apply is based on self-learning. The model is initially trained using a limited number of labeled examples. Then confident predictions for unlabelled examples are added to the set of labeled examples for retraining of the model. Generally, multiple iterations of labelling and retraining are performed. One important hyper-parameter is the selection percentage $p$, which specifies how many of the most confident predictions are added to the training set after each iteration. We use a small value of $p$ in our experiments to select the most confident predictions only. Generating many more labeled data points in this fashion allows for deep neural networks to be trained to their full capacity, and generally results in significant performance improvements. For more details on this approach see, for instance, our previous work in~\cite{attaullah2019ssl}.

 In this paper, for network weight initialisation, we transfer pretrained weights from Imagenet classification and fine-tune on the target domain. We compare the performance achieved by this weight transfer to the performance of training using a fully random initialisation of the weights of the neural network. 
 
The proposed approach is very general, suggesting that a spectrum of loss functions and label propagation algorithms can all work well in this framework. We use the most widely used classification loss, i.e., softmax cross-entropy, as a first option. In addition, we explore loss functions based on similarity metric learning. The embeddings produced by the neural network after training with a similarity function can be employed to assign class labels to unlabeled examples using very simple classification methods such as a nearest-neighbor classifier.  Below we review the loss functions used for the experiments.

\subsection{Softmax Cross-entropy Loss}
The single most frequently used classification loss function is softmax cross-entropy, which is a measure of the difference between the desired probability distribution and the predicted probability distribution:
 
\begin{equation}
\label{cross-entropy}
\mathcal{L} = - \frac{1}{N}\sum\limits_{i = 1}^N {\log } \frac{{{e^{W_{{y_i}}^T{x_i} + {b_{{y_i}}}}}}}{{\sum\limits_{j = 1}^c {{e^{W_j^T{x_i} + {b_j}}}} }},\tag{1}
\end{equation}

where $ x_i \in \mathbb{R}^d$ denotes the deep features (the "embedding") of the $i^{th}$ sample, belonging to the class $y_i$, and $d$ is the dimension of the embedding,  $W_j \in \mathbb{R}^d$ denotes the $j^{th}$ column of the weight matrix $W \in \mathbb{R}^{d \times c}$ and $ b_j  \in \mathbb{R}^c$ is the bias term. The batch size for gradient descent is $ N $ and $ c $ is the number of classes.

\subsection{Siamese Networks}
Siamese networks~\cite{bromley1993signature} are neural networks for training a similarity function given labeled data using one of several possible loss functions. They can be thought of as two identical copies of the same network, sharing all weights. They are particularly suitable for datasets with many classes containing only a few labeled instances per class and can employ any of the loss functions listed below.

\subsubsection{Triplet Loss}
The triplet loss~\cite{schroff2015facenet} is widely used. A triplet's anchor example $ a $, positive example $ p $, and negative example $ n $  are provided as a training example to the network for getting corresponding embeddings. Normally $a$ and $p$ come from the same class, and $n$ is from a different class. Triplet loss tries to push the negative example's embedding farther away from positive example's one, with a user-specified minimum margin $m$. Using, e.g., Euclidean distance $d(.,.)$ between embedded examples, the triplet loss is calculated as:

\begin{equation}
\label{triplet-loss}
\mathcal{L} = max(d(a, p) - d(a, n) + m, 0)
\end{equation}

Triplet loss tries to push $d(a,p)$ to 0 and $d(a,n)$ to be greater than $d(a,p)+m$. Triplets can be categorized as:
\begin{itemize}
	\item \textbf{Easy triplets}: those with a loss of 0.
	\item \textbf{Hard triplets}: those where  $n$ is closer to  $a$ than  $p$.
	\item \textbf{Semi-hard triplets}: those where  $n$ is not closer to  $a$ than  $p$, but is within the margin, thus still returning a positive loss.
\end{itemize}
In our experiments, we use semi-hard triplets for training of the neural network as they yield more distinctive embeddings \cite{schroff2015facenet}.

\subsubsection{Contrastive loss}
The contrastive loss~\cite{hadsell2006dimensionality} is a pair-based loss that attempts to bring similar examples closer to each other and push dissimilar examples farther away with respect to a minimum margin $ m $. Contrastive loss for embeddings of two examples $x_1$ and $x_2$ can be calculated as follows:

\begin{equation}
\label{contrastive}
\mathcal{L} = y \times d(x_1 , x_2) + (1-y) \times max(0,m - d(x_1 , x_2)) 
\end{equation}
Here, $y=1$ if $x_1$ and $x_2$ are from the same class, and $y=0$ otherwise.

\subsubsection{ArcFace loss}
Arcface loss~\cite{deng2019arcface} is a modified cross-entropy loss with angular margins in
the softmax expression, which is designed for improved discrimination in metric learning. The loss is calculated as:

\begin{equation}
\label{arcface-loss}
\mathcal{L} = - \frac{1}{N}\sum\limits_{i = 1}^N {\log } \frac{{{e^{s\left( {\cos \left( {{\theta _{{y_i}}} + m} \right)} \right)}}}}{{{e^{s\left( {\cos \left( {{\theta _{{y_i}}} + m} \right)} \right)}} + \sum {_{j = 1,j \ne {y_i}}^c{e^{s\cos {\theta _j}}}} }}.\tag{3}
\end{equation}

$ \theta_j $ is the angle between the $ l_2 $-normalized weight vector $ W_j $ and the feature vector $ x_i $. The bias term $ b_j $ is ignored for simplicity. The feature vector $ x_i $ is $ l_2 $-normalised and scaled to $ s $, the radius of the hypersphere. An additive angular margin penalty $ m $ 
is added to the ground truth angle $ \theta_{y_i} $.

\section{Experiments}
For evaluating the effect of transfer learning, we consider three image classification problems. For all datasets, a small subset of labeled examples was chosen according to standard semi-supervised learning practice, with a balanced number of examples from each class. All remaining examples were used as unlabeled training examples. For triplet, contrastive and Arcface loss, $k$-nearest neighbor is used for label prediction, with $k=1$ for simplicity\footnote{Source code is available at \url{https://github.com/attaullah/Self-training/blob/master/Transfer_learning.md}}. 
We always include two network version in the comparison: one using randomly initialised weights, and one using pretrained weights from ImageNet.
All models are evaluated on the standard test split for each dataset in three different ways: after training only on the initially labeled examples, then after training for a number of meta-iterations using our semi-supervised learning approaches, and also --- for comparison --- after training on all labeled training examples. 
The two sets of results computed from a) only the initial labeled examples, and b)  all labeled training examples, act as an empirical lower and upper bound for the semi-supervised approaches.

We used the VGG16 network architecture for all experiments. A fully connected layer is added at the end of the model for generating a 256-dimensional embedding space. A mini-batch size of 100 is used for all the experiments. For updating the network parameters, Adam is used as the optimizer, except for contrastive loss, which uses Rmsprop. For triplet, contrastive, and Arcface loss, the distance to the nearest labeled example is used as the confidence score when selecting unlabeled examples for labeling. For softmax cross-entropy loss, the softmax probability score is used as the confidence score. Our proposed self-learning approach was run for 25 meta-iterations and results were averaged over 3 runs with a random selection of initially labeled examples.

\subsection{Results}
SVHN (Street View House Numbers) comprises 32x32 color images of house numbers. A single image can contain  multiple digits, but only the digit in the center is considered for the label prediction. The proposed approaches are evaluated using 1000 labeled instances initially and use a selection percentage of $5\%$  (i.e., in each meta-iteration of self-training, 5\% of the remaining unlabeled examples are selected for labeling). Table \ref{tabel:svhn} shows test accuracy for SVHN using all four losses, with random as well as pretrained weights, for the 1000-labeled, the self-learning, and the all-labeled setup.

\begin{table}
	\caption{SVHN   Test Accuracy \%.}
	\label{tabel:svhn}
	\begin{center}
	\begin{tabular}{|l|r|r|r|}
		\hline
		Pretrained         & {1000 Labels} & {Self-learning} & {73257 Labels}  \\
		\hline     
		\multicolumn{4}{|l|}{Cross-entropy loss} \\
		\hline
		No  & 75.81  $\pm$ 2.28 & 92.07 $\pm$ 0.35 & 95.72 $\pm$  0.23   \\
		Yes  & 80.84  $\pm$ 0.74     &  \textbf{92.73 $\pm$ 0.52 }& \textbf{96.10 $\pm$  0.21}                          \\
		\hline
		
		\multicolumn{4}{|l|}{ Triplet loss~\cite{schroff2015facenet} } \\
		\hline
		 No          & 57.22  $\pm$ 1.81    &64.69 $\pm$ 1.39  & 94.79 $\pm$  0.06   \\
		 Yes         &  \textbf{82.52  $\pm$ 2.14}      & 86.14 $\pm$ 1.11 &  95.12 $\pm$  0.23   \\

 		\hline
		 \multicolumn{4}{|l|}{Contrastive loss~\cite{hadsell2006dimensionality}} \\
		 \hline
		 No		&	54.73  $\pm$ 0.57 	  &	62.80 $\pm$ 0.63 	& 81.82 $\pm$  2.29		\\
		 Yes 	&79.46  $\pm$ 0.99	 	  &	82.59 $\pm$ 0.31   & 93.41 $\pm$  0.26   \\

		\hline     
		\multicolumn{4}{|l|}{Arcface loss~\cite{deng2019arcface}} \\
		\hline
		No  &  68.33  $\pm$ 0.91&70.42 $\pm$ 1.59 & 93.74 $\pm$  0.11  \\
		Yes &80.84  $\pm$ 0.21  & 82.01 $\pm$ 1.41  & 95.66 $\pm$  0.31               \\
		\hline     
	\end{tabular}
\end{center}
\end{table}

The CIFAR-10 dataset comprises 32x32 RGB images of ten different object classes. The proposed semi-supervised approaches are evaluated using 4000 labeled instances initially, with a selection percentage of $5\%$ for self-training.  Table \ref{tabel:cifar10} shows accuracy on the standard test set for all losses using 4000-labeled, all-labeled and self-learning, for pretrained weights from Imagenet as well as random initial weights. 

\begin{table}
	\caption{CIFAR10   Test Accuracy \%.}
	\label{tabel:cifar10}
	\begin{center}
		\begin{tabular}{|l|r|r|r|}
			\hline
			Pretrained         & {4000 Labels} & {Self-learning} & {50000 Labels}  \\
			\hline
			\multicolumn{4}{|l|}{Cross-entropy loss} \\
			\hline
			No  &70.43 $\pm$ 1.43      & 79.15 $\pm$ 0.80 & 87.84 $\pm$  0.39    \\
			Yes         &\textbf{77.07  $\pm$ 0.91}       & \textbf{83.33 $\pm$ 0.19} & \textbf{89.37 $\pm$  0.49}   \\

			\hline  
			\multicolumn{4}{|l|}{ Triplet loss~\cite{schroff2015facenet} } \\
			\hline
			No          &68.35 $\pm$ 3.63  & 70.57 $\pm$ 1.17 & 86.54 $\pm$ 0.42   \\
			Yes         & 76.42 $\pm$ 2.19      & 78.36 $\pm$ 1.39 & 88.15 $\pm$ 0.36   \\

			\hline
			\multicolumn{4}{|l|}{Contrastive loss~\cite{hadsell2006dimensionality}} \\
			\hline
			No		&34.90  $\pm$ 0.73      & 44.58 $\pm$ 1.67  &71.16 $\pm$  0.05    \\
			Yes 	&71.98  $\pm$ 0.95  & 76.58 $\pm$ 0.05 & 85.92 $\pm$  0.32    \\
			
			\hline     
			   
			\multicolumn{4}{|l|}{Arcface loss~\cite{deng2019arcface}} \\
			\hline
			No  &  55.04  $\pm$ 1.36      &69.54 $\pm$ 3.69  &75.31 $\pm$  0.24   \\
			Yes &74.76  $\pm$ 0.72 &76.55$\pm$ 1.80  &87.76 $\pm$  0.24     \\	
			\hline     
		\end{tabular}
	\end{center}
\end{table}

\begin{figure}
	\centering
	\includegraphics[scale=0.53] {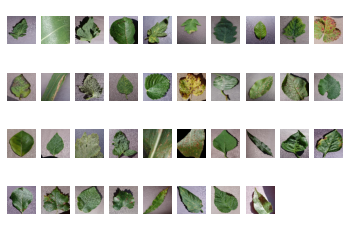}
	\caption{Plant Village disease~\cite{hughes2015open} dataset}
	\label{fig:plant-dataset}
\end{figure}

The Plant Village~\cite{hughes2015open}  dataset consists of plant leaves\footnote{Dataset is available at \url{https://github.com/attaullah/downsampled-plant-disease-dataset}}. It has 43,456 training and 10,849 test RGB images resized to 96x96 from the original format (256x256). It has 38 categories of species and diseases. A sample image for each class is shown in Figure~\ref{fig:plant-dataset}. The proposed semi-supervised approaches are evaluated using 10 images per class as labeled instances initially, with a selection percentage of $2\%$ in self-learning. 
Table \ref{tabel:plant} shows accuracy on test examples for all four losses using 380-labeled, all-labeled and self-learning, with random weight initialization and pretrained weights.

\begin{table}
	\caption{Plant village 96x96 Test Accuracy \%.}
	\label{tabel:plant}
	\begin{center}
		\begin{tabular}{|l|r|r|r|}
			\hline
			Pretrained         & {380 Labels} & {Self-learning} & {43456 Labels}  \\
			\hline
			\multicolumn{4}{|l|}{Cross-entropy loss} \\
			\hline
			No  &45.78  $\pm$ 4.09      & 54.58 $\pm$ 2.65 & 98.24 $\pm$  0.62    \\
			Yes         &73.76  $\pm$ 1.70       & \textbf{84.62 $\pm$ 1.2} & 99.24 $\pm$  0.08   \\
			\hline  
			\multicolumn{4}{|l|}{ Triplet loss~\cite{schroff2015facenet} } \\
			\hline
			No          &29.81  $\pm$ 2.59 & 33.16 $\pm$ 1.96  & 92.15 $\pm$  1.63    \\
			Yes         & \textbf{76.88  $\pm$ 0.36}   & 77.80 $\pm$ 1.15 &99.02 $\pm$  0.11    \\
			\hline
			\multicolumn{4}{|l|}{Contrastive loss~\cite{hadsell2006dimensionality}} \\
			\hline
			No		& 13.12  $\pm$ 1.56    & 16.35 $\pm$ 0.88 &  34.75 $\pm$  3.20   \\
			Yes 	& 30.22  $\pm$ 2.14  & 32.46 $\pm$ 2.65& 45.66 $\pm$  2.64     \\
			
			\hline     
			   
			\multicolumn{4}{|l|}{Arcface loss~\cite{deng2019arcface}} \\
			\hline
			No  & 54.85  $\pm$ 0.09    & 58.39 $\pm$ 3.61 &98.11 $\pm$  0.38    \\
			Yes & 60.67  $\pm$ 0.13& 71.80 $\pm$ 2.58  & \textbf{99.32 $\pm$  0.04}    \\	
			\hline     
		\end{tabular}
	\end{center}
\end{table}
As we can see from the results for all three datasets, using pretrained weights  generally results in substantial improvements over random initialisation. When comparing the four loss functions, cross-entropy emerges as the winner, with triplet loss often being second best. However, especially for small numbers of labeled examples, triplet loss seems competitive with cross-entropy, outperforming it for two of the three datasets. This seems reasonable, as paying explicit attention to the similarities of particular instances may be more important when only a few labeled instances are available. 

Comparing the three metric losses with each other, triplet loss generally outperforms the other two when using pretrained weights. On the other hand, when using random initial weights, none of the three losses seems to have a clear advantage over the others, except for the Plant dataset, where Arcface performs very well, even outperforming cross-entropy.

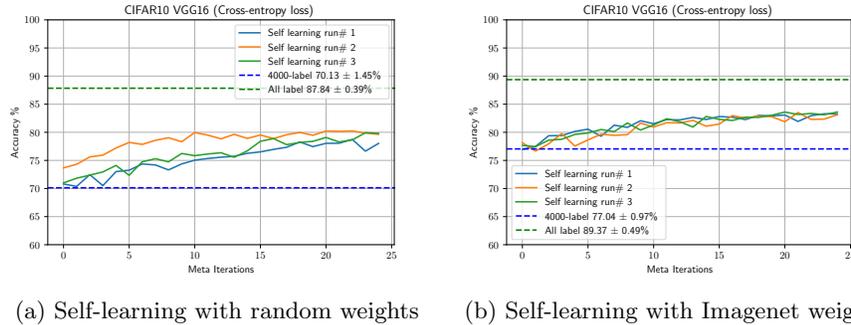
\begin{figure}
	\centering 
	\begin{subfigure}{0.5\textwidth}
		\centering
	\scalebox{0.39} {\input{imgs/random_weights.pgf} }
		\caption{Self-learning with random weights}
		\label{fig:cifar10-random}
	\end{subfigure}\hfil 
	\begin{subfigure}{0.5\textwidth}
		\centering	
		\scalebox{0.39} {\input{imgs/pretrained_weights.pgf} }
		\caption{Self-learning with Imagenet weights}
		\label{fig:cifar10-pretrained}
	\end{subfigure} 
	\caption{CIFAR10 meta-iterations of self-learning using random and pretrained Imagenet weights.}
	\label{fig:cifar10randomvspretrained}
\end{figure}

Figure \ref{fig:cifar10randomvspretrained} shows a comparison of self-learning using random weights and pretrained weights, across three different runs on CIFAR10, using softmax cross-entropy loss for 4000 initially labeled examples and 25 meta-iterations of self-learning.  The accuracy curves show similar improvements for both scenarios, with the pretrained version starting from a higher initial accuracy level, and retaining this advantage over the 25 meta-iterations of self-learning.

\begin{figure}
	\centering 
	\begin{subfigure}{0.5\textwidth}
		\centering
		\includegraphics[scale=0.167] {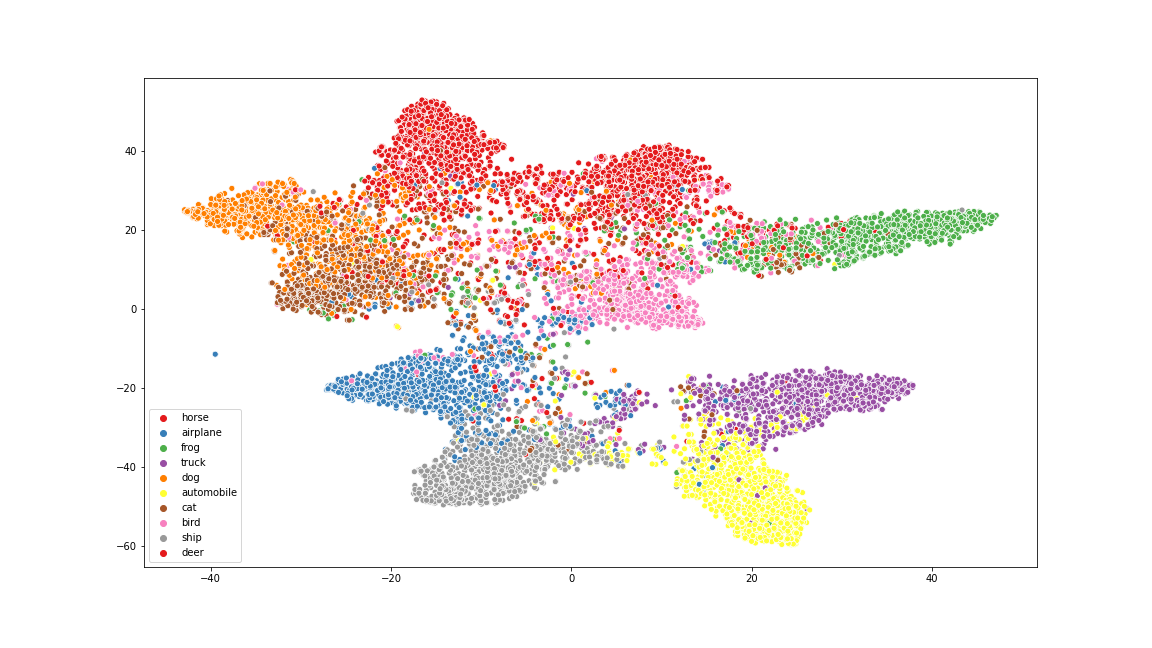}
		\caption{Cross-entropy: initial training }
		\label{fig:cross-entropy-4k}
	\end{subfigure}\hfil 
	\begin{subfigure}{0.5\textwidth}
		\includegraphics[scale=0.167] {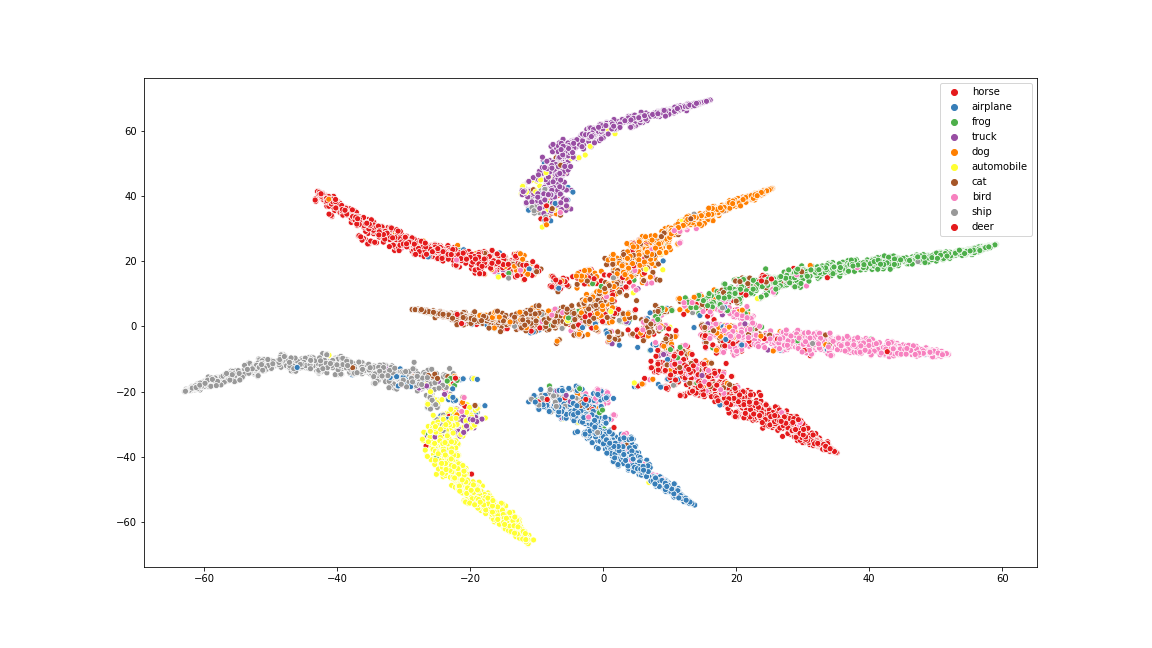}
		\caption{Cross-entropy: after 25 meta-iterations}
		\label{fig:cross-entropy-25meta}
	\end{subfigure}

	\medskip
	
	\begin{subfigure}{0.5\textwidth}
		\centering
		\includegraphics[scale=0.167] {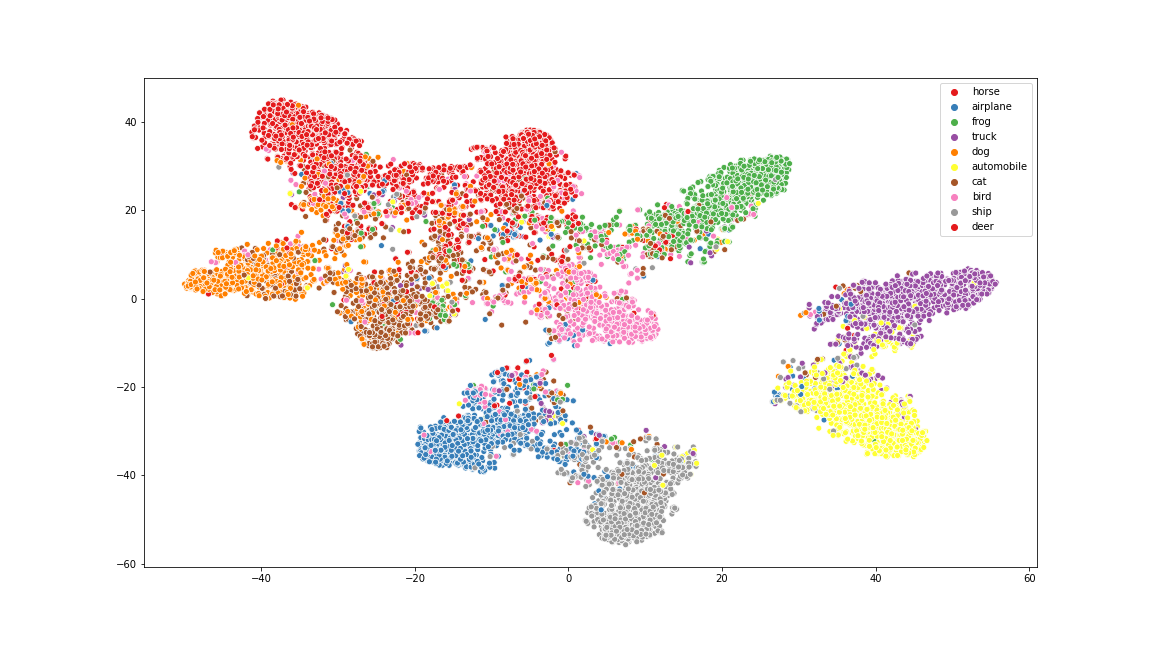}
		\caption{Triplet loss: initial training }
		\label{fig:triplet-4k}
	\end{subfigure}\hfil 
	\begin{subfigure}{0.5\textwidth}
		\centering
		\includegraphics[scale=0.167] {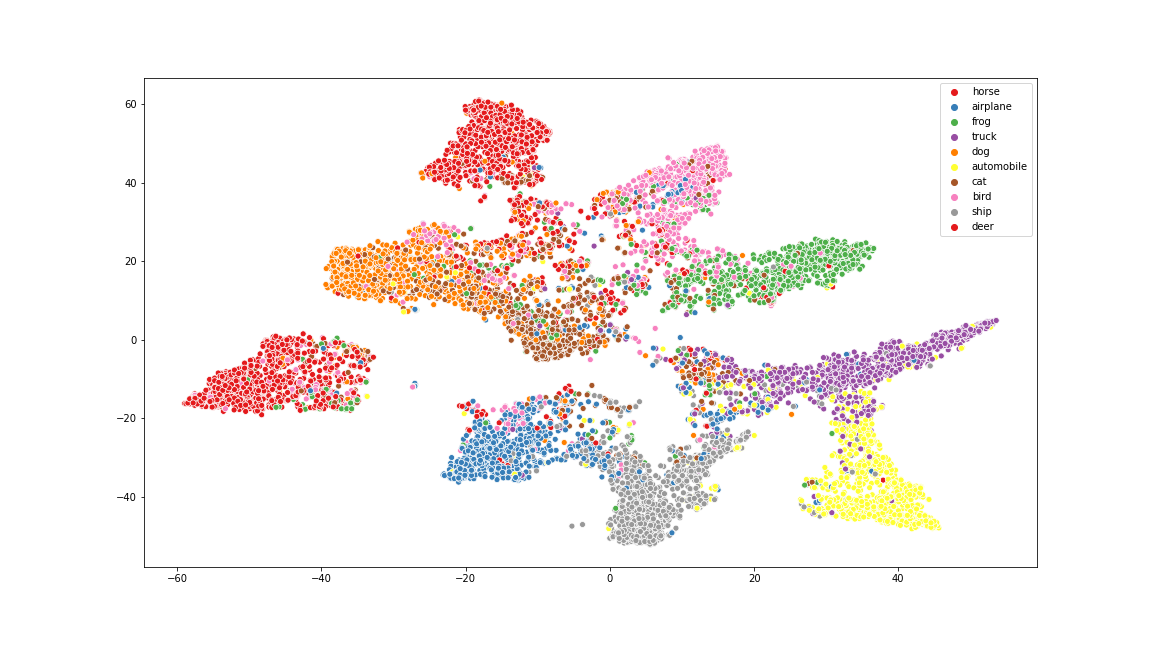}
		\caption{Triplet loss: after 25 meta-iterations}
		\label{fig:triplet-25meta}
	\end{subfigure} 
	
	\medskip
	
	\begin{subfigure}{0.5\textwidth}
		\centering
		\includegraphics[scale=0.167] {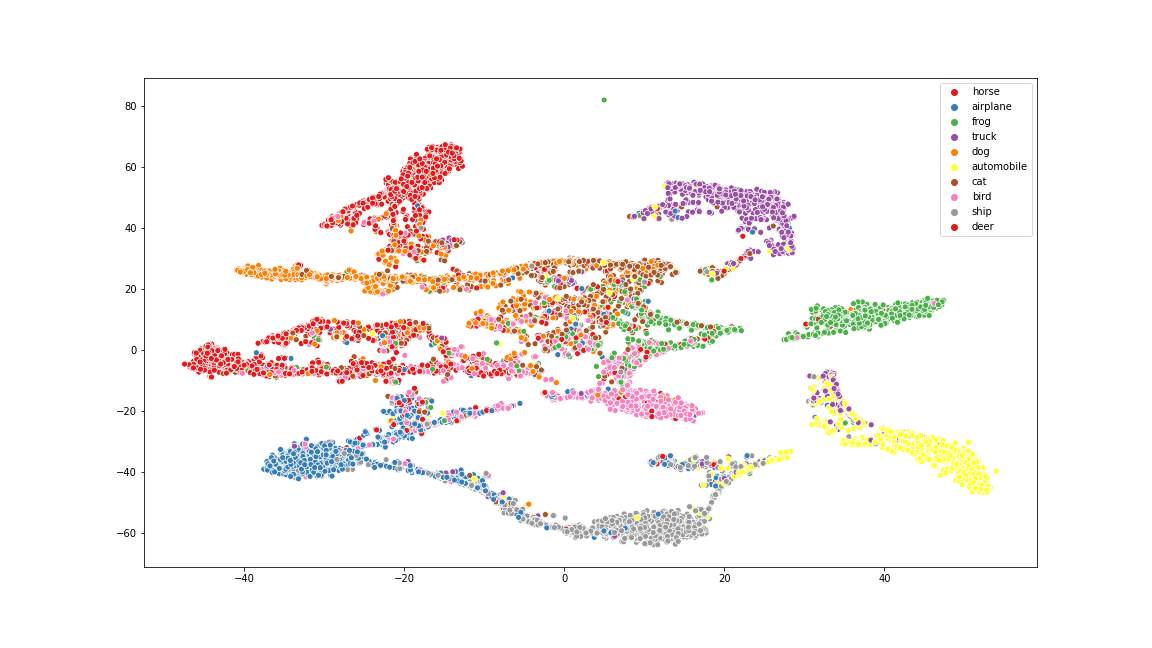} 
		\caption{Contrastive loss: initial training }
		\label{fig:contrastive-4k}
	\end{subfigure}\hfil 
	\begin{subfigure}{0.5\textwidth}
		\includegraphics[scale=0.167] {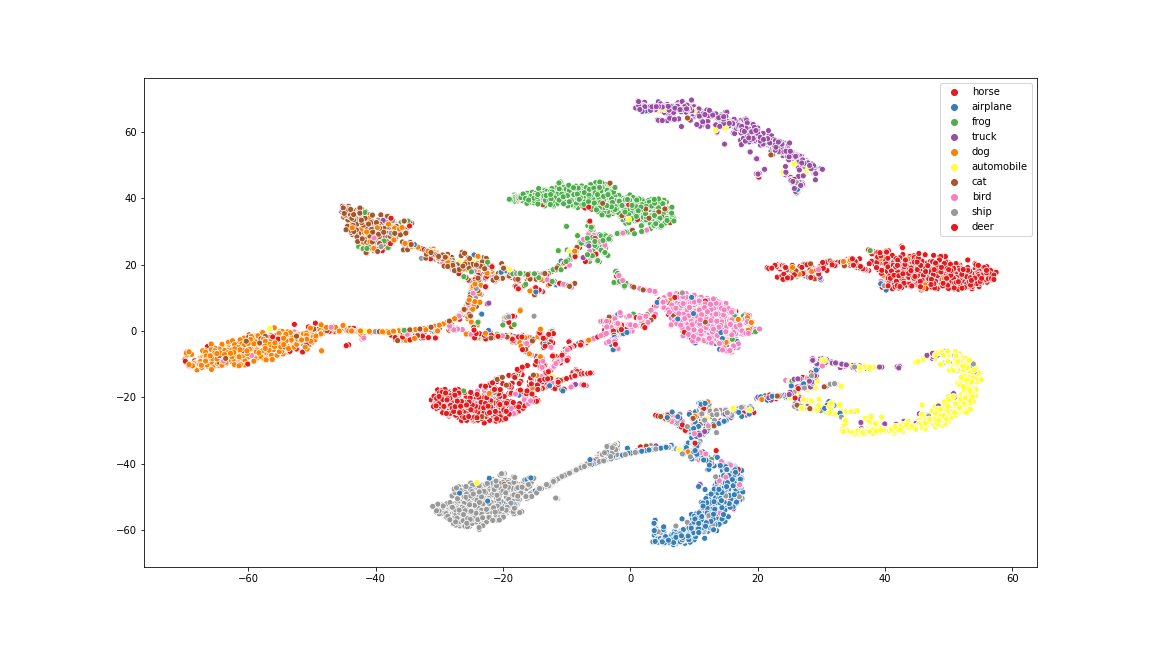} 
		\caption{Contrastive loss: after 25 meta-iterations}
		\label{fig:contrastive-25meta}
	\end{subfigure}

	\medskip

	\begin{subfigure}{0.5\textwidth}
		\centering
		\includegraphics[scale=0.167] {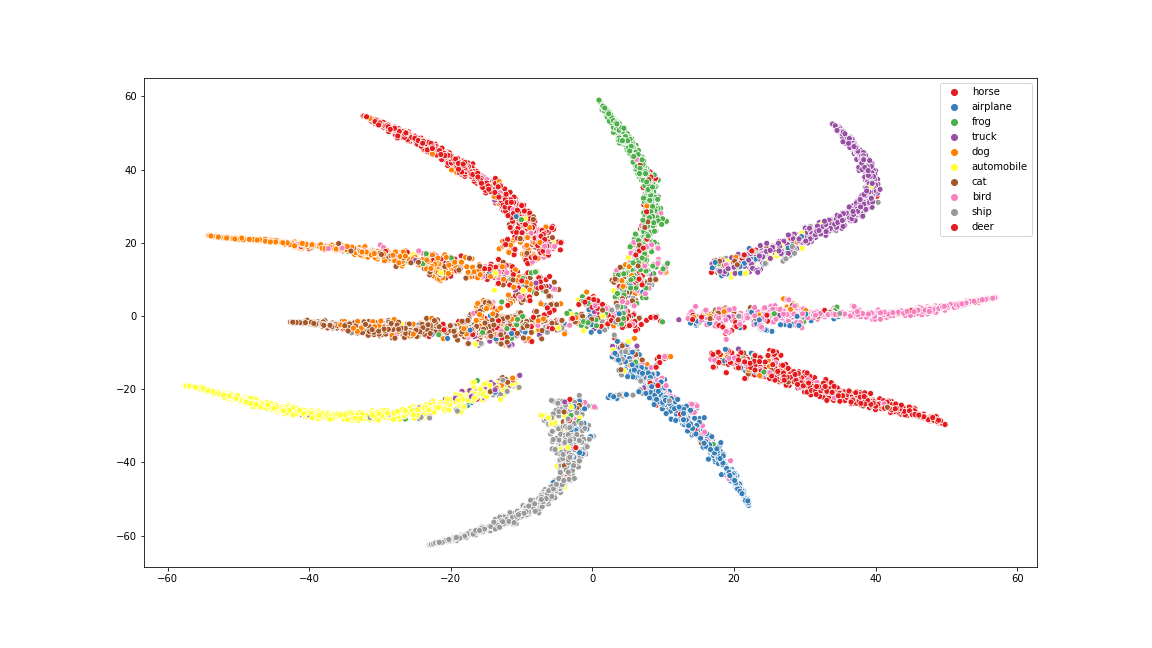}
		\caption{Arcface: initial training}
		\label{fig:arcface-4k}
	\end{subfigure}\hfil 
	\begin{subfigure}{0.5\textwidth}
		\includegraphics[scale=0.167] {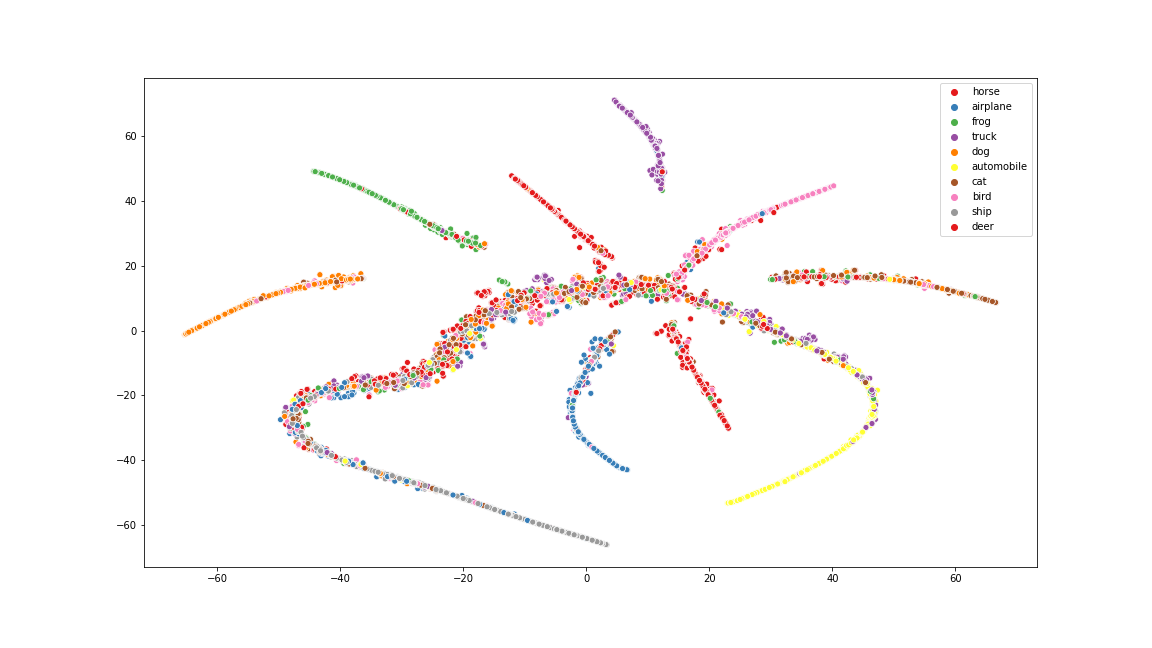}
		\caption{Arcface: after 25 meta-iterations}
		\label{fig:arcface-25meta}
	\end{subfigure}
\caption{TSNE Visualization of CIFAR10 embeddings for all losses after the first 4000 labeled examples and after 25-meta iterations of self-learning.}
	\label{fig:embeddingsall}
\end{figure}

In order to investigate the effect of self-learning on the embeddings, we visualize the embeddings  obtained using all four loss functions. Figure \ref{fig:embeddingsall} shows the output of TSNE~\cite{van2008visualizing} on embeddings of CIFAR10 test instances after training on 4000 labeled examples and after 25 meta-iterations of self-learning using all four losses. It is evident that self-learning improves class separation, with cross-entropy showing the most dramatic improvement, consistent with its high final accuracy.

\section{Conclusions}
In this	paper, we have shown that transfer learning can be highly beneficial for semi-supervised image classification. In	terms of loss functions, overall, cross-entropy outperforms more specialised losses like triplet loss, contrastive loss, or Arcface loss. Still, for a small number of labels, triplet loss is  very competitive.

There are a number of directions for future work. Exploring combinations	of well-performing loss	functions, exploring alternatives to	the label propagation scheme, and exploring connections to few-shot	learning, are just a few	obvious	ones. Additionally, more lower-level engineering ideas, like mini-batch composition strategies as pointed out in~\cite{arazo2020pseudo}, might help to further improve	the performance	of semi-supervised image classification.

\bibliographystyle{splncs04}
\bibliography{References}

\end{document}

%% file: imgs/random_weights.pgf
\begingroup%
\makeatletter%
\begin{pgfpicture}%
\pgfpathrectangle{\pgfpointorigin}{\pgfqpoint{6.000000in}{4.000000in}}%
\pgfusepath{use as bounding box, clip}%
\begin{pgfscope}%
\pgfsetbuttcap%
\pgfsetmiterjoin%
\definecolor{currentfill}{rgb}{1.000000,1.000000,1.000000}%
\pgfsetfillcolor{currentfill}%
\pgfsetlinewidth{0.000000pt}%
\definecolor{currentstroke}{rgb}{1.000000,1.000000,1.000000}%
\pgfsetstrokecolor{currentstroke}%
\pgfsetdash{}{0pt}%
\pgfpathmoveto{\pgfqpoint{0.000000in}{0.000000in}}%
\pgfpathlineto{\pgfqpoint{6.000000in}{0.000000in}}%
\pgfpathlineto{\pgfqpoint{6.000000in}{4.000000in}}%
\pgfpathlineto{\pgfqpoint{0.000000in}{4.000000in}}%
\pgfpathclose%
\pgfusepath{fill}%
\end{pgfscope}%
\begin{pgfscope}%
\pgfsetbuttcap%
\pgfsetmiterjoin%
\definecolor{currentfill}{rgb}{1.000000,1.000000,1.000000}%
\pgfsetfillcolor{currentfill}%
\pgfsetlinewidth{0.000000pt}%
\definecolor{currentstroke}{rgb}{0.000000,0.000000,0.000000}%
\pgfsetstrokecolor{currentstroke}%
\pgfsetstrokeopacity{0.000000}%
\pgfsetdash{}{0pt}%
\pgfpathmoveto{\pgfqpoint{0.750000in}{0.500000in}}%
\pgfpathlineto{\pgfqpoint{5.400000in}{0.500000in}}%
\pgfpathlineto{\pgfqpoint{5.400000in}{3.520000in}}%
\pgfpathlineto{\pgfqpoint{0.750000in}{3.520000in}}%
\pgfpathclose%
\pgfusepath{fill}%
\end{pgfscope}%
\begin{pgfscope}%
\pgfpathrectangle{\pgfqpoint{0.750000in}{0.500000in}}{\pgfqpoint{4.650000in}{3.020000in}}%
\pgfusepath{clip}%
\pgfsetrectcap%
\pgfsetroundjoin%
\pgfsetlinewidth{0.803000pt}%
\definecolor{currentstroke}{rgb}{0.690196,0.690196,0.690196}%
\pgfsetstrokecolor{currentstroke}%
\pgfsetdash{}{0pt}%
\pgfpathmoveto{\pgfqpoint{0.961364in}{0.500000in}}%
\pgfpathlineto{\pgfqpoint{0.961364in}{3.520000in}}%
\pgfusepath{stroke}%
\end{pgfscope}%
\begin{pgfscope}%
\pgfsetbuttcap%
\pgfsetroundjoin%
\definecolor{currentfill}{rgb}{0.000000,0.000000,0.000000}%
\pgfsetfillcolor{currentfill}%
\pgfsetlinewidth{0.803000pt}%
\definecolor{currentstroke}{rgb}{0.000000,0.000000,0.000000}%
\pgfsetstrokecolor{currentstroke}%
\pgfsetdash{}{0pt}%
\pgfsys@defobject{currentmarker}{\pgfqpoint{0.000000in}{-0.048611in}}{\pgfqpoint{0.000000in}{0.000000in}}{%
\pgfpathmoveto{\pgfqpoint{0.000000in}{0.000000in}}%
\pgfpathlineto{\pgfqpoint{0.000000in}{-0.048611in}}%
\pgfusepath{stroke,fill}%
}%
\begin{pgfscope}%
\pgfsys@transformshift{0.961364in}{0.500000in}%
\pgfsys@useobject{currentmarker}{}%
\end{pgfscope}%
\end{pgfscope}%
\begin{pgfscope}%
\definecolor{textcolor}{rgb}{0.000000,0.000000,0.000000}%
\pgfsetstrokecolor{textcolor}%
\pgfsetfillcolor{textcolor}%
\pgftext[x=0.961364in,y=0.402778in,,top]{\color{textcolor}\sffamily\fontsize{10.000000}{12.000000}\selectfont \(\displaystyle {0}\)}%
\end{pgfscope}%
\begin{pgfscope}%
\pgfpathrectangle{\pgfqpoint{0.750000in}{0.500000in}}{\pgfqpoint{4.650000in}{3.020000in}}%
\pgfusepath{clip}%
\pgfsetrectcap%
\pgfsetroundjoin%
\pgfsetlinewidth{0.803000pt}%
\definecolor{currentstroke}{rgb}{0.690196,0.690196,0.690196}%
\pgfsetstrokecolor{currentstroke}%
\pgfsetdash{}{0pt}%
\pgfpathmoveto{\pgfqpoint{1.842045in}{0.500000in}}%
\pgfpathlineto{\pgfqpoint{1.842045in}{3.520000in}}%
\pgfusepath{stroke}%
\end{pgfscope}%
\begin{pgfscope}%
\pgfsetbuttcap%
\pgfsetroundjoin%
\definecolor{currentfill}{rgb}{0.000000,0.000000,0.000000}%
\pgfsetfillcolor{currentfill}%
\pgfsetlinewidth{0.803000pt}%
\definecolor{currentstroke}{rgb}{0.000000,0.000000,0.000000}%
\pgfsetstrokecolor{currentstroke}%
\pgfsetdash{}{0pt}%
\pgfsys@defobject{currentmarker}{\pgfqpoint{0.000000in}{-0.048611in}}{\pgfqpoint{0.000000in}{0.000000in}}{%
\pgfpathmoveto{\pgfqpoint{0.000000in}{0.000000in}}%
\pgfpathlineto{\pgfqpoint{0.000000in}{-0.048611in}}%
\pgfusepath{stroke,fill}%
}%
\begin{pgfscope}%
\pgfsys@transformshift{1.842045in}{0.500000in}%
\pgfsys@useobject{currentmarker}{}%
\end{pgfscope}%
\end{pgfscope}%
\begin{pgfscope}%
\definecolor{textcolor}{rgb}{0.000000,0.000000,0.000000}%
\pgfsetstrokecolor{textcolor}%
\pgfsetfillcolor{textcolor}%
\pgftext[x=1.842045in,y=0.402778in,,top]{\color{textcolor}\sffamily\fontsize{10.000000}{12.000000}\selectfont \(\displaystyle {5}\)}%
\end{pgfscope}%
\begin{pgfscope}%
\pgfpathrectangle{\pgfqpoint{0.750000in}{0.500000in}}{\pgfqpoint{4.650000in}{3.020000in}}%
\pgfusepath{clip}%
\pgfsetrectcap%
\pgfsetroundjoin%
\pgfsetlinewidth{0.803000pt}%
\definecolor{currentstroke}{rgb}{0.690196,0.690196,0.690196}%
\pgfsetstrokecolor{currentstroke}%
\pgfsetdash{}{0pt}%
\pgfpathmoveto{\pgfqpoint{2.722727in}{0.500000in}}%
\pgfpathlineto{\pgfqpoint{2.722727in}{3.520000in}}%
\pgfusepath{stroke}%
\end{pgfscope}%
\begin{pgfscope}%
\pgfsetbuttcap%
\pgfsetroundjoin%
\definecolor{currentfill}{rgb}{0.000000,0.000000,0.000000}%
\pgfsetfillcolor{currentfill}%
\pgfsetlinewidth{0.803000pt}%
\definecolor{currentstroke}{rgb}{0.000000,0.000000,0.000000}%
\pgfsetstrokecolor{currentstroke}%
\pgfsetdash{}{0pt}%
\pgfsys@defobject{currentmarker}{\pgfqpoint{0.000000in}{-0.048611in}}{\pgfqpoint{0.000000in}{0.000000in}}{%
\pgfpathmoveto{\pgfqpoint{0.000000in}{0.000000in}}%
\pgfpathlineto{\pgfqpoint{0.000000in}{-0.048611in}}%
\pgfusepath{stroke,fill}%
}%
\begin{pgfscope}%
\pgfsys@transformshift{2.722727in}{0.500000in}%
\pgfsys@useobject{currentmarker}{}%
\end{pgfscope}%
\end{pgfscope}%
\begin{pgfscope}%
\definecolor{textcolor}{rgb}{0.000000,0.000000,0.000000}%
\pgfsetstrokecolor{textcolor}%
\pgfsetfillcolor{textcolor}%
\pgftext[x=2.722727in,y=0.402778in,,top]{\color{textcolor}\sffamily\fontsize{10.000000}{12.000000}\selectfont \(\displaystyle {10}\)}%
\end{pgfscope}%
\begin{pgfscope}%
\pgfpathrectangle{\pgfqpoint{0.750000in}{0.500000in}}{\pgfqpoint{4.650000in}{3.020000in}}%
\pgfusepath{clip}%
\pgfsetrectcap%
\pgfsetroundjoin%
\pgfsetlinewidth{0.803000pt}%
\definecolor{currentstroke}{rgb}{0.690196,0.690196,0.690196}%
\pgfsetstrokecolor{currentstroke}%
\pgfsetdash{}{0pt}%
\pgfpathmoveto{\pgfqpoint{3.603409in}{0.500000in}}%
\pgfpathlineto{\pgfqpoint{3.603409in}{3.520000in}}%
\pgfusepath{stroke}%
\end{pgfscope}%
\begin{pgfscope}%
\pgfsetbuttcap%
\pgfsetroundjoin%
\definecolor{currentfill}{rgb}{0.000000,0.000000,0.000000}%
\pgfsetfillcolor{currentfill}%
\pgfsetlinewidth{0.803000pt}%
\definecolor{currentstroke}{rgb}{0.000000,0.000000,0.000000}%
\pgfsetstrokecolor{currentstroke}%
\pgfsetdash{}{0pt}%
\pgfsys@defobject{currentmarker}{\pgfqpoint{0.000000in}{-0.048611in}}{\pgfqpoint{0.000000in}{0.000000in}}{%
\pgfpathmoveto{\pgfqpoint{0.000000in}{0.000000in}}%
\pgfpathlineto{\pgfqpoint{0.000000in}{-0.048611in}}%
\pgfusepath{stroke,fill}%
}%
\begin{pgfscope}%
\pgfsys@transformshift{3.603409in}{0.500000in}%
\pgfsys@useobject{currentmarker}{}%
\end{pgfscope}%
\end{pgfscope}%
\begin{pgfscope}%
\definecolor{textcolor}{rgb}{0.000000,0.000000,0.000000}%
\pgfsetstrokecolor{textcolor}%
\pgfsetfillcolor{textcolor}%
\pgftext[x=3.603409in,y=0.402778in,,top]{\color{textcolor}\sffamily\fontsize{10.000000}{12.000000}\selectfont \(\displaystyle {15}\)}%
\end{pgfscope}%
\begin{pgfscope}%
\pgfpathrectangle{\pgfqpoint{0.750000in}{0.500000in}}{\pgfqpoint{4.650000in}{3.020000in}}%
\pgfusepath{clip}%
\pgfsetrectcap%
\pgfsetroundjoin%
\pgfsetlinewidth{0.803000pt}%
\definecolor{currentstroke}{rgb}{0.690196,0.690196,0.690196}%
\pgfsetstrokecolor{currentstroke}%
\pgfsetdash{}{0pt}%
\pgfpathmoveto{\pgfqpoint{4.484091in}{0.500000in}}%
\pgfpathlineto{\pgfqpoint{4.484091in}{3.520000in}}%
\pgfusepath{stroke}%
\end{pgfscope}%
\begin{pgfscope}%
\pgfsetbuttcap%
\pgfsetroundjoin%
\definecolor{currentfill}{rgb}{0.000000,0.000000,0.000000}%
\pgfsetfillcolor{currentfill}%
\pgfsetlinewidth{0.803000pt}%
\definecolor{currentstroke}{rgb}{0.000000,0.000000,0.000000}%
\pgfsetstrokecolor{currentstroke}%
\pgfsetdash{}{0pt}%
\pgfsys@defobject{currentmarker}{\pgfqpoint{0.000000in}{-0.048611in}}{\pgfqpoint{0.000000in}{0.000000in}}{%
\pgfpathmoveto{\pgfqpoint{0.000000in}{0.000000in}}%
\pgfpathlineto{\pgfqpoint{0.000000in}{-0.048611in}}%
\pgfusepath{stroke,fill}%
}%
\begin{pgfscope}%
\pgfsys@transformshift{4.484091in}{0.500000in}%
\pgfsys@useobject{currentmarker}{}%
\end{pgfscope}%
\end{pgfscope}%
\begin{pgfscope}%
\definecolor{textcolor}{rgb}{0.000000,0.000000,0.000000}%
\pgfsetstrokecolor{textcolor}%
\pgfsetfillcolor{textcolor}%
\pgftext[x=4.484091in,y=0.402778in,,top]{\color{textcolor}\sffamily\fontsize{10.000000}{12.000000}\selectfont \(\displaystyle {20}\)}%
\end{pgfscope}%
\begin{pgfscope}%
\pgfpathrectangle{\pgfqpoint{0.750000in}{0.500000in}}{\pgfqpoint{4.650000in}{3.020000in}}%
\pgfusepath{clip}%
\pgfsetrectcap%
\pgfsetroundjoin%
\pgfsetlinewidth{0.803000pt}%
\definecolor{currentstroke}{rgb}{0.690196,0.690196,0.690196}%
\pgfsetstrokecolor{currentstroke}%
\pgfsetdash{}{0pt}%
\pgfpathmoveto{\pgfqpoint{5.364773in}{0.500000in}}%
\pgfpathlineto{\pgfqpoint{5.364773in}{3.520000in}}%
\pgfusepath{stroke}%
\end{pgfscope}%
\begin{pgfscope}%
\pgfsetbuttcap%
\pgfsetroundjoin%
\definecolor{currentfill}{rgb}{0.000000,0.000000,0.000000}%
\pgfsetfillcolor{currentfill}%
\pgfsetlinewidth{0.803000pt}%
\definecolor{currentstroke}{rgb}{0.000000,0.000000,0.000000}%
\pgfsetstrokecolor{currentstroke}%
\pgfsetdash{}{0pt}%
\pgfsys@defobject{currentmarker}{\pgfqpoint{0.000000in}{-0.048611in}}{\pgfqpoint{0.000000in}{0.000000in}}{%
\pgfpathmoveto{\pgfqpoint{0.000000in}{0.000000in}}%
\pgfpathlineto{\pgfqpoint{0.000000in}{-0.048611in}}%
\pgfusepath{stroke,fill}%
}%
\begin{pgfscope}%
\pgfsys@transformshift{5.364773in}{0.500000in}%
\pgfsys@useobject{currentmarker}{}%
\end{pgfscope}%
\end{pgfscope}%
\begin{pgfscope}%
\definecolor{textcolor}{rgb}{0.000000,0.000000,0.000000}%
\pgfsetstrokecolor{textcolor}%
\pgfsetfillcolor{textcolor}%
\pgftext[x=5.364773in,y=0.402778in,,top]{\color{textcolor}\sffamily\fontsize{10.000000}{12.000000}\selectfont \(\displaystyle {25}\)}%
\end{pgfscope}%
\begin{pgfscope}%
\definecolor{textcolor}{rgb}{0.000000,0.000000,0.000000}%
\pgfsetstrokecolor{textcolor}%
\pgfsetfillcolor{textcolor}%
\pgftext[x=3.075000in,y=0.227624in,,top]{\color{textcolor}\sffamily\fontsize{10.000000}{12.000000}\selectfont Meta Iterations}%
\end{pgfscope}%
\begin{pgfscope}%
\pgfpathrectangle{\pgfqpoint{0.750000in}{0.500000in}}{\pgfqpoint{4.650000in}{3.020000in}}%
\pgfusepath{clip}%
\pgfsetrectcap%
\pgfsetroundjoin%
\pgfsetlinewidth{0.803000pt}%
\definecolor{currentstroke}{rgb}{0.690196,0.690196,0.690196}%
\pgfsetstrokecolor{currentstroke}%
\pgfsetdash{}{0pt}%
\pgfpathmoveto{\pgfqpoint{0.750000in}{0.500000in}}%
\pgfpathlineto{\pgfqpoint{5.400000in}{0.500000in}}%
\pgfusepath{stroke}%
\end{pgfscope}%
\begin{pgfscope}%
\pgfsetbuttcap%
\pgfsetroundjoin%
\definecolor{currentfill}{rgb}{0.000000,0.000000,0.000000}%
\pgfsetfillcolor{currentfill}%
\pgfsetlinewidth{0.803000pt}%
\definecolor{currentstroke}{rgb}{0.000000,0.000000,0.000000}%
\pgfsetstrokecolor{currentstroke}%
\pgfsetdash{}{0pt}%
\pgfsys@defobject{currentmarker}{\pgfqpoint{-0.048611in}{0.000000in}}{\pgfqpoint{0.000000in}{0.000000in}}{%
\pgfpathmoveto{\pgfqpoint{0.000000in}{0.000000in}}%
\pgfpathlineto{\pgfqpoint{-0.048611in}{0.000000in}}%
\pgfusepath{stroke,fill}%
}%
\begin{pgfscope}%
\pgfsys@transformshift{0.750000in}{0.500000in}%
\pgfsys@useobject{currentmarker}{}%
\end{pgfscope}%
\end{pgfscope}%
\begin{pgfscope}%
\definecolor{textcolor}{rgb}{0.000000,0.000000,0.000000}%
\pgfsetstrokecolor{textcolor}%
\pgfsetfillcolor{textcolor}%
\pgftext[x=0.506946in, y=0.451775in, left, base]{\color{textcolor}\sffamily\fontsize{10.000000}{12.000000}\selectfont \(\displaystyle {60}\)}%
\end{pgfscope}%
\begin{pgfscope}%
\pgfpathrectangle{\pgfqpoint{0.750000in}{0.500000in}}{\pgfqpoint{4.650000in}{3.020000in}}%
\pgfusepath{clip}%
\pgfsetrectcap%
\pgfsetroundjoin%
\pgfsetlinewidth{0.803000pt}%
\definecolor{currentstroke}{rgb}{0.690196,0.690196,0.690196}%
\pgfsetstrokecolor{currentstroke}%
\pgfsetdash{}{0pt}%
\pgfpathmoveto{\pgfqpoint{0.750000in}{0.877500in}}%
\pgfpathlineto{\pgfqpoint{5.400000in}{0.877500in}}%
\pgfusepath{stroke}%
\end{pgfscope}%
\begin{pgfscope}%
\pgfsetbuttcap%
\pgfsetroundjoin%
\definecolor{currentfill}{rgb}{0.000000,0.000000,0.000000}%
\pgfsetfillcolor{currentfill}%
\pgfsetlinewidth{0.803000pt}%
\definecolor{currentstroke}{rgb}{0.000000,0.000000,0.000000}%
\pgfsetstrokecolor{currentstroke}%
\pgfsetdash{}{0pt}%
\pgfsys@defobject{currentmarker}{\pgfqpoint{-0.048611in}{0.000000in}}{\pgfqpoint{0.000000in}{0.000000in}}{%
\pgfpathmoveto{\pgfqpoint{0.000000in}{0.000000in}}%
\pgfpathlineto{\pgfqpoint{-0.048611in}{0.000000in}}%
\pgfusepath{stroke,fill}%
}%
\begin{pgfscope}%
\pgfsys@transformshift{0.750000in}{0.877500in}%
\pgfsys@useobject{currentmarker}{}%
\end{pgfscope}%
\end{pgfscope}%
\begin{pgfscope}%
\definecolor{textcolor}{rgb}{0.000000,0.000000,0.000000}%
\pgfsetstrokecolor{textcolor}%
\pgfsetfillcolor{textcolor}%
\pgftext[x=0.506946in, y=0.829275in, left, base]{\color{textcolor}\sffamily\fontsize{10.000000}{12.000000}\selectfont \(\displaystyle {65}\)}%
\end{pgfscope}%
\begin{pgfscope}%
\pgfpathrectangle{\pgfqpoint{0.750000in}{0.500000in}}{\pgfqpoint{4.650000in}{3.020000in}}%
\pgfusepath{clip}%
\pgfsetrectcap%
\pgfsetroundjoin%
\pgfsetlinewidth{0.803000pt}%
\definecolor{currentstroke}{rgb}{0.690196,0.690196,0.690196}%
\pgfsetstrokecolor{currentstroke}%
\pgfsetdash{}{0pt}%
\pgfpathmoveto{\pgfqpoint{0.750000in}{1.255000in}}%
\pgfpathlineto{\pgfqpoint{5.400000in}{1.255000in}}%
\pgfusepath{stroke}%
\end{pgfscope}%
\begin{pgfscope}%
\pgfsetbuttcap%
\pgfsetroundjoin%
\definecolor{currentfill}{rgb}{0.000000,0.000000,0.000000}%
\pgfsetfillcolor{currentfill}%
\pgfsetlinewidth{0.803000pt}%
\definecolor{currentstroke}{rgb}{0.000000,0.000000,0.000000}%
\pgfsetstrokecolor{currentstroke}%
\pgfsetdash{}{0pt}%
\pgfsys@defobject{currentmarker}{\pgfqpoint{-0.048611in}{0.000000in}}{\pgfqpoint{0.000000in}{0.000000in}}{%
\pgfpathmoveto{\pgfqpoint{0.000000in}{0.000000in}}%
\pgfpathlineto{\pgfqpoint{-0.048611in}{0.000000in}}%
\pgfusepath{stroke,fill}%
}%
\begin{pgfscope}%
\pgfsys@transformshift{0.750000in}{1.255000in}%
\pgfsys@useobject{currentmarker}{}%
\end{pgfscope}%
\end{pgfscope}%
\begin{pgfscope}%
\definecolor{textcolor}{rgb}{0.000000,0.000000,0.000000}%
\pgfsetstrokecolor{textcolor}%
\pgfsetfillcolor{textcolor}%
\pgftext[x=0.506946in, y=1.206775in, left, base]{\color{textcolor}\sffamily\fontsize{10.000000}{12.000000}\selectfont \(\displaystyle {70}\)}%
\end{pgfscope}%
\begin{pgfscope}%
\pgfpathrectangle{\pgfqpoint{0.750000in}{0.500000in}}{\pgfqpoint{4.650000in}{3.020000in}}%
\pgfusepath{clip}%
\pgfsetrectcap%
\pgfsetroundjoin%
\pgfsetlinewidth{0.803000pt}%
\definecolor{currentstroke}{rgb}{0.690196,0.690196,0.690196}%
\pgfsetstrokecolor{currentstroke}%
\pgfsetdash{}{0pt}%
\pgfpathmoveto{\pgfqpoint{0.750000in}{1.632500in}}%
\pgfpathlineto{\pgfqpoint{5.400000in}{1.632500in}}%
\pgfusepath{stroke}%
\end{pgfscope}%
\begin{pgfscope}%
\pgfsetbuttcap%
\pgfsetroundjoin%
\definecolor{currentfill}{rgb}{0.000000,0.000000,0.000000}%
\pgfsetfillcolor{currentfill}%
\pgfsetlinewidth{0.803000pt}%
\definecolor{currentstroke}{rgb}{0.000000,0.000000,0.000000}%
\pgfsetstrokecolor{currentstroke}%
\pgfsetdash{}{0pt}%
\pgfsys@defobject{currentmarker}{\pgfqpoint{-0.048611in}{0.000000in}}{\pgfqpoint{0.000000in}{0.000000in}}{%
\pgfpathmoveto{\pgfqpoint{0.000000in}{0.000000in}}%
\pgfpathlineto{\pgfqpoint{-0.048611in}{0.000000in}}%
\pgfusepath{stroke,fill}%
}%
\begin{pgfscope}%
\pgfsys@transformshift{0.750000in}{1.632500in}%
\pgfsys@useobject{currentmarker}{}%
\end{pgfscope}%
\end{pgfscope}%
\begin{pgfscope}%
\definecolor{textcolor}{rgb}{0.000000,0.000000,0.000000}%
\pgfsetstrokecolor{textcolor}%
\pgfsetfillcolor{textcolor}%
\pgftext[x=0.506946in, y=1.584275in, left, base]{\color{textcolor}\sffamily\fontsize{10.000000}{12.000000}\selectfont \(\displaystyle {75}\)}%
\end{pgfscope}%
\begin{pgfscope}%
\pgfpathrectangle{\pgfqpoint{0.750000in}{0.500000in}}{\pgfqpoint{4.650000in}{3.020000in}}%
\pgfusepath{clip}%
\pgfsetrectcap%
\pgfsetroundjoin%
\pgfsetlinewidth{0.803000pt}%
\definecolor{currentstroke}{rgb}{0.690196,0.690196,0.690196}%
\pgfsetstrokecolor{currentstroke}%
\pgfsetdash{}{0pt}%
\pgfpathmoveto{\pgfqpoint{0.750000in}{2.010000in}}%
\pgfpathlineto{\pgfqpoint{5.400000in}{2.010000in}}%
\pgfusepath{stroke}%
\end{pgfscope}%
\begin{pgfscope}%
\pgfsetbuttcap%
\pgfsetroundjoin%
\definecolor{currentfill}{rgb}{0.000000,0.000000,0.000000}%
\pgfsetfillcolor{currentfill}%
\pgfsetlinewidth{0.803000pt}%
\definecolor{currentstroke}{rgb}{0.000000,0.000000,0.000000}%
\pgfsetstrokecolor{currentstroke}%
\pgfsetdash{}{0pt}%
\pgfsys@defobject{currentmarker}{\pgfqpoint{-0.048611in}{0.000000in}}{\pgfqpoint{0.000000in}{0.000000in}}{%
\pgfpathmoveto{\pgfqpoint{0.000000in}{0.000000in}}%
\pgfpathlineto{\pgfqpoint{-0.048611in}{0.000000in}}%
\pgfusepath{stroke,fill}%
}%
\begin{pgfscope}%
\pgfsys@transformshift{0.750000in}{2.010000in}%
\pgfsys@useobject{currentmarker}{}%
\end{pgfscope}%
\end{pgfscope}%
\begin{pgfscope}%
\definecolor{textcolor}{rgb}{0.000000,0.000000,0.000000}%
\pgfsetstrokecolor{textcolor}%
\pgfsetfillcolor{textcolor}%
\pgftext[x=0.506946in, y=1.961775in, left, base]{\color{textcolor}\sffamily\fontsize{10.000000}{12.000000}\selectfont \(\displaystyle {80}\)}%
\end{pgfscope}%
\begin{pgfscope}%
\pgfpathrectangle{\pgfqpoint{0.750000in}{0.500000in}}{\pgfqpoint{4.650000in}{3.020000in}}%
\pgfusepath{clip}%
\pgfsetrectcap%
\pgfsetroundjoin%
\pgfsetlinewidth{0.803000pt}%
\definecolor{currentstroke}{rgb}{0.690196,0.690196,0.690196}%
\pgfsetstrokecolor{currentstroke}%
\pgfsetdash{}{0pt}%
\pgfpathmoveto{\pgfqpoint{0.750000in}{2.387500in}}%
\pgfpathlineto{\pgfqpoint{5.400000in}{2.387500in}}%
\pgfusepath{stroke}%
\end{pgfscope}%
\begin{pgfscope}%
\pgfsetbuttcap%
\pgfsetroundjoin%
\definecolor{currentfill}{rgb}{0.000000,0.000000,0.000000}%
\pgfsetfillcolor{currentfill}%
\pgfsetlinewidth{0.803000pt}%
\definecolor{currentstroke}{rgb}{0.000000,0.000000,0.000000}%
\pgfsetstrokecolor{currentstroke}%
\pgfsetdash{}{0pt}%
\pgfsys@defobject{currentmarker}{\pgfqpoint{-0.048611in}{0.000000in}}{\pgfqpoint{0.000000in}{0.000000in}}{%
\pgfpathmoveto{\pgfqpoint{0.000000in}{0.000000in}}%
\pgfpathlineto{\pgfqpoint{-0.048611in}{0.000000in}}%
\pgfusepath{stroke,fill}%
}%
\begin{pgfscope}%
\pgfsys@transformshift{0.750000in}{2.387500in}%
\pgfsys@useobject{currentmarker}{}%
\end{pgfscope}%
\end{pgfscope}%
\begin{pgfscope}%
\definecolor{textcolor}{rgb}{0.000000,0.000000,0.000000}%
\pgfsetstrokecolor{textcolor}%
\pgfsetfillcolor{textcolor}%
\pgftext[x=0.506946in, y=2.339275in, left, base]{\color{textcolor}\sffamily\fontsize{10.000000}{12.000000}\selectfont \(\displaystyle {85}\)}%
\end{pgfscope}%
\begin{pgfscope}%
\pgfpathrectangle{\pgfqpoint{0.750000in}{0.500000in}}{\pgfqpoint{4.650000in}{3.020000in}}%
\pgfusepath{clip}%
\pgfsetrectcap%
\pgfsetroundjoin%
\pgfsetlinewidth{0.803000pt}%
\definecolor{currentstroke}{rgb}{0.690196,0.690196,0.690196}%
\pgfsetstrokecolor{currentstroke}%
\pgfsetdash{}{0pt}%
\pgfpathmoveto{\pgfqpoint{0.750000in}{2.765000in}}%
\pgfpathlineto{\pgfqpoint{5.400000in}{2.765000in}}%
\pgfusepath{stroke}%
\end{pgfscope}%
\begin{pgfscope}%
\pgfsetbuttcap%
\pgfsetroundjoin%
\definecolor{currentfill}{rgb}{0.000000,0.000000,0.000000}%
\pgfsetfillcolor{currentfill}%
\pgfsetlinewidth{0.803000pt}%
\definecolor{currentstroke}{rgb}{0.000000,0.000000,0.000000}%
\pgfsetstrokecolor{currentstroke}%
\pgfsetdash{}{0pt}%
\pgfsys@defobject{currentmarker}{\pgfqpoint{-0.048611in}{0.000000in}}{\pgfqpoint{0.000000in}{0.000000in}}{%
\pgfpathmoveto{\pgfqpoint{0.000000in}{0.000000in}}%
\pgfpathlineto{\pgfqpoint{-0.048611in}{0.000000in}}%
\pgfusepath{stroke,fill}%
}%
\begin{pgfscope}%
\pgfsys@transformshift{0.750000in}{2.765000in}%
\pgfsys@useobject{currentmarker}{}%
\end{pgfscope}%
\end{pgfscope}%
\begin{pgfscope}%
\definecolor{textcolor}{rgb}{0.000000,0.000000,0.000000}%
\pgfsetstrokecolor{textcolor}%
\pgfsetfillcolor{textcolor}%
\pgftext[x=0.506946in, y=2.716775in, left, base]{\color{textcolor}\sffamily\fontsize{10.000000}{12.000000}\selectfont \(\displaystyle {90}\)}%
\end{pgfscope}%
\begin{pgfscope}%
\pgfpathrectangle{\pgfqpoint{0.750000in}{0.500000in}}{\pgfqpoint{4.650000in}{3.020000in}}%
\pgfusepath{clip}%
\pgfsetrectcap%
\pgfsetroundjoin%
\pgfsetlinewidth{0.803000pt}%
\definecolor{currentstroke}{rgb}{0.690196,0.690196,0.690196}%
\pgfsetstrokecolor{currentstroke}%
\pgfsetdash{}{0pt}%
\pgfpathmoveto{\pgfqpoint{0.750000in}{3.142500in}}%
\pgfpathlineto{\pgfqpoint{5.400000in}{3.142500in}}%
\pgfusepath{stroke}%
\end{pgfscope}%
\begin{pgfscope}%
\pgfsetbuttcap%
\pgfsetroundjoin%
\definecolor{currentfill}{rgb}{0.000000,0.000000,0.000000}%
\pgfsetfillcolor{currentfill}%
\pgfsetlinewidth{0.803000pt}%
\definecolor{currentstroke}{rgb}{0.000000,0.000000,0.000000}%
\pgfsetstrokecolor{currentstroke}%
\pgfsetdash{}{0pt}%
\pgfsys@defobject{currentmarker}{\pgfqpoint{-0.048611in}{0.000000in}}{\pgfqpoint{0.000000in}{0.000000in}}{%
\pgfpathmoveto{\pgfqpoint{0.000000in}{0.000000in}}%
\pgfpathlineto{\pgfqpoint{-0.048611in}{0.000000in}}%
\pgfusepath{stroke,fill}%
}%
\begin{pgfscope}%
\pgfsys@transformshift{0.750000in}{3.142500in}%
\pgfsys@useobject{currentmarker}{}%
\end{pgfscope}%
\end{pgfscope}%
\begin{pgfscope}%
\definecolor{textcolor}{rgb}{0.000000,0.000000,0.000000}%
\pgfsetstrokecolor{textcolor}%
\pgfsetfillcolor{textcolor}%
\pgftext[x=0.506946in, y=3.094275in, left, base]{\color{textcolor}\sffamily\fontsize{10.000000}{12.000000}\selectfont \(\displaystyle {95}\)}%
\end{pgfscope}%
\begin{pgfscope}%
\pgfpathrectangle{\pgfqpoint{0.750000in}{0.500000in}}{\pgfqpoint{4.650000in}{3.020000in}}%
\pgfusepath{clip}%
\pgfsetrectcap%
\pgfsetroundjoin%
\pgfsetlinewidth{0.803000pt}%
\definecolor{currentstroke}{rgb}{0.690196,0.690196,0.690196}%
\pgfsetstrokecolor{currentstroke}%
\pgfsetdash{}{0pt}%
\pgfpathmoveto{\pgfqpoint{0.750000in}{3.520000in}}%
\pgfpathlineto{\pgfqpoint{5.400000in}{3.520000in}}%
\pgfusepath{stroke}%
\end{pgfscope}%
\begin{pgfscope}%
\pgfsetbuttcap%
\pgfsetroundjoin%
\definecolor{currentfill}{rgb}{0.000000,0.000000,0.000000}%
\pgfsetfillcolor{currentfill}%
\pgfsetlinewidth{0.803000pt}%
\definecolor{currentstroke}{rgb}{0.000000,0.000000,0.000000}%
\pgfsetstrokecolor{currentstroke}%
\pgfsetdash{}{0pt}%
\pgfsys@defobject{currentmarker}{\pgfqpoint{-0.048611in}{0.000000in}}{\pgfqpoint{0.000000in}{0.000000in}}{%
\pgfpathmoveto{\pgfqpoint{0.000000in}{0.000000in}}%
\pgfpathlineto{\pgfqpoint{-0.048611in}{0.000000in}}%
\pgfusepath{stroke,fill}%
}%
\begin{pgfscope}%
\pgfsys@transformshift{0.750000in}{3.520000in}%
\pgfsys@useobject{currentmarker}{}%
\end{pgfscope}%
\end{pgfscope}%
\begin{pgfscope}%
\definecolor{textcolor}{rgb}{0.000000,0.000000,0.000000}%
\pgfsetstrokecolor{textcolor}%
\pgfsetfillcolor{textcolor}%
\pgftext[x=0.434030in, y=3.471775in, left, base]{\color{textcolor}\sffamily\fontsize{10.000000}{12.000000}\selectfont \(\displaystyle {100}\)}%
\end{pgfscope}%
\begin{pgfscope}%
\definecolor{textcolor}{rgb}{0.000000,0.000000,0.000000}%
\pgfsetstrokecolor{textcolor}%
\pgfsetfillcolor{textcolor}%
\pgftext[x=0.378474in,y=2.010000in,,bottom,rotate=90.000000]{\color{textcolor}\sffamily\fontsize{10.000000}{12.000000}\selectfont Accuracy \%}%
\end{pgfscope}%
\begin{pgfscope}%
\pgfpathrectangle{\pgfqpoint{0.750000in}{0.500000in}}{\pgfqpoint{4.650000in}{3.020000in}}%
\pgfusepath{clip}%
\pgfsetrectcap%
\pgfsetroundjoin%
\pgfsetlinewidth{1.505625pt}%
\definecolor{currentstroke}{rgb}{0.121569,0.466667,0.705882}%
\pgfsetstrokecolor{currentstroke}%
\pgfsetdash{}{0pt}%
\pgfpathmoveto{\pgfqpoint{0.961364in}{1.315400in}}%
\pgfpathlineto{\pgfqpoint{1.137500in}{1.283690in}}%
\pgfpathlineto{\pgfqpoint{1.313636in}{1.441485in}}%
\pgfpathlineto{\pgfqpoint{1.489773in}{1.293505in}}%
\pgfpathlineto{\pgfqpoint{1.665909in}{1.482255in}}%
\pgfpathlineto{\pgfqpoint{1.842045in}{1.500375in}}%
\pgfpathlineto{\pgfqpoint{2.018182in}{1.584935in}}%
\pgfpathlineto{\pgfqpoint{2.194318in}{1.571345in}}%
\pgfpathlineto{\pgfqpoint{2.370455in}{1.504905in}}%
\pgfpathlineto{\pgfqpoint{2.546591in}{1.584935in}}%
\pgfpathlineto{\pgfqpoint{2.722727in}{1.635520in}}%
\pgfpathlineto{\pgfqpoint{2.898864in}{1.658170in}}%
\pgfpathlineto{\pgfqpoint{3.075000in}{1.675535in}}%
\pgfpathlineto{\pgfqpoint{3.251136in}{1.687615in}}%
\pgfpathlineto{\pgfqpoint{3.427273in}{1.727630in}}%
\pgfpathlineto{\pgfqpoint{3.603409in}{1.747260in}}%
\pgfpathlineto{\pgfqpoint{3.779545in}{1.780480in}}%
\pgfpathlineto{\pgfqpoint{3.955682in}{1.809170in}}%
\pgfpathlineto{\pgfqpoint{4.131818in}{1.880140in}}%
\pgfpathlineto{\pgfqpoint{4.307955in}{1.817475in}}%
\pgfpathlineto{\pgfqpoint{4.484091in}{1.862020in}}%
\pgfpathlineto{\pgfqpoint{4.660227in}{1.862775in}}%
\pgfpathlineto{\pgfqpoint{4.836364in}{1.914870in}}%
\pgfpathlineto{\pgfqpoint{5.012500in}{1.756320in}}%
\pgfpathlineto{\pgfqpoint{5.188636in}{1.860510in}}%
\pgfusepath{stroke}%
\end{pgfscope}%
\begin{pgfscope}%
\pgfpathrectangle{\pgfqpoint{0.750000in}{0.500000in}}{\pgfqpoint{4.650000in}{3.020000in}}%
\pgfusepath{clip}%
\pgfsetrectcap%
\pgfsetroundjoin%
\pgfsetlinewidth{1.505625pt}%
\definecolor{currentstroke}{rgb}{1.000000,0.498039,0.054902}%
\pgfsetstrokecolor{currentstroke}%
\pgfsetdash{}{0pt}%
\pgfpathmoveto{\pgfqpoint{0.961364in}{1.532085in}}%
\pgfpathlineto{\pgfqpoint{1.137500in}{1.580405in}}%
\pgfpathlineto{\pgfqpoint{1.313636in}{1.679310in}}%
\pgfpathlineto{\pgfqpoint{1.489773in}{1.703470in}}%
\pgfpathlineto{\pgfqpoint{1.665909in}{1.799355in}}%
\pgfpathlineto{\pgfqpoint{1.842045in}{1.874100in}}%
\pgfpathlineto{\pgfqpoint{2.018182in}{1.847675in}}%
\pgfpathlineto{\pgfqpoint{2.194318in}{1.902035in}}%
\pgfpathlineto{\pgfqpoint{2.370455in}{1.936765in}}%
\pgfpathlineto{\pgfqpoint{2.546591in}{1.883160in}}%
\pgfpathlineto{\pgfqpoint{2.722727in}{2.006980in}}%
\pgfpathlineto{\pgfqpoint{2.898864in}{1.971495in}}%
\pgfpathlineto{\pgfqpoint{3.075000in}{1.922420in}}%
\pgfpathlineto{\pgfqpoint{3.251136in}{1.982820in}}%
\pgfpathlineto{\pgfqpoint{3.427273in}{1.927705in}}%
\pgfpathlineto{\pgfqpoint{3.603409in}{1.973760in}}%
\pgfpathlineto{\pgfqpoint{3.779545in}{1.925440in}}%
\pgfpathlineto{\pgfqpoint{3.955682in}{1.977535in}}%
\pgfpathlineto{\pgfqpoint{4.131818in}{2.007735in}}%
\pgfpathlineto{\pgfqpoint{4.307955in}{1.969985in}}%
\pgfpathlineto{\pgfqpoint{4.484091in}{2.027365in}}%
\pgfpathlineto{\pgfqpoint{4.660227in}{2.022835in}}%
\pgfpathlineto{\pgfqpoint{4.836364in}{2.028120in}}%
\pgfpathlineto{\pgfqpoint{5.012500in}{2.000185in}}%
\pgfpathlineto{\pgfqpoint{5.188636in}{1.982820in}}%
\pgfusepath{stroke}%
\end{pgfscope}%
\begin{pgfscope}%
\pgfpathrectangle{\pgfqpoint{0.750000in}{0.500000in}}{\pgfqpoint{4.650000in}{3.020000in}}%
\pgfusepath{clip}%
\pgfsetrectcap%
\pgfsetroundjoin%
\pgfsetlinewidth{1.505625pt}%
\definecolor{currentstroke}{rgb}{0.172549,0.627451,0.172549}%
\pgfsetstrokecolor{currentstroke}%
\pgfsetdash{}{0pt}%
\pgfpathmoveto{\pgfqpoint{0.961364in}{1.331255in}}%
\pgfpathlineto{\pgfqpoint{1.137500in}{1.393165in}}%
\pgfpathlineto{\pgfqpoint{1.313636in}{1.434690in}}%
\pgfpathlineto{\pgfqpoint{1.489773in}{1.477725in}}%
\pgfpathlineto{\pgfqpoint{1.665909in}{1.565305in}}%
\pgfpathlineto{\pgfqpoint{1.842045in}{1.432425in}}%
\pgfpathlineto{\pgfqpoint{2.018182in}{1.613625in}}%
\pgfpathlineto{\pgfqpoint{2.194318in}{1.655150in}}%
\pgfpathlineto{\pgfqpoint{2.370455in}{1.612870in}}%
\pgfpathlineto{\pgfqpoint{2.546591in}{1.725365in}}%
\pgfpathlineto{\pgfqpoint{2.722727in}{1.695920in}}%
\pgfpathlineto{\pgfqpoint{2.898864in}{1.718570in}}%
\pgfpathlineto{\pgfqpoint{3.075000in}{1.735180in}}%
\pgfpathlineto{\pgfqpoint{3.251136in}{1.674780in}}%
\pgfpathlineto{\pgfqpoint{3.427273in}{1.760095in}}%
\pgfpathlineto{\pgfqpoint{3.603409in}{1.889200in}}%
\pgfpathlineto{\pgfqpoint{3.779545in}{1.925440in}}%
\pgfpathlineto{\pgfqpoint{3.955682in}{1.843900in}}%
\pgfpathlineto{\pgfqpoint{4.131818in}{1.873345in}}%
\pgfpathlineto{\pgfqpoint{4.307955in}{1.888445in}}%
\pgfpathlineto{\pgfqpoint{4.484091in}{1.940540in}}%
\pgfpathlineto{\pgfqpoint{4.660227in}{1.880140in}}%
\pgfpathlineto{\pgfqpoint{4.836364in}{1.907320in}}%
\pgfpathlineto{\pgfqpoint{5.012500in}{2.004715in}}%
\pgfpathlineto{\pgfqpoint{5.188636in}{1.995655in}}%
\pgfusepath{stroke}%
\end{pgfscope}%
\begin{pgfscope}%
\pgfpathrectangle{\pgfqpoint{0.750000in}{0.500000in}}{\pgfqpoint{4.650000in}{3.020000in}}%
\pgfusepath{clip}%
\pgfsetbuttcap%
\pgfsetroundjoin%
\pgfsetlinewidth{1.505625pt}%
\definecolor{currentstroke}{rgb}{0.000000,0.000000,1.000000}%
\pgfsetstrokecolor{currentstroke}%
\pgfsetdash{{5.550000pt}{2.400000pt}}{0.000000pt}%
\pgfpathmoveto{\pgfqpoint{0.750000in}{1.264815in}}%
\pgfpathlineto{\pgfqpoint{5.413889in}{1.264815in}}%
\pgfusepath{stroke}%
\end{pgfscope}%
\begin{pgfscope}%
\pgfpathrectangle{\pgfqpoint{0.750000in}{0.500000in}}{\pgfqpoint{4.650000in}{3.020000in}}%
\pgfusepath{clip}%
\pgfsetbuttcap%
\pgfsetroundjoin%
\pgfsetlinewidth{1.505625pt}%
\definecolor{currentstroke}{rgb}{0.000000,0.500000,0.000000}%
\pgfsetstrokecolor{currentstroke}%
\pgfsetdash{{5.550000pt}{2.400000pt}}{0.000000pt}%
\pgfpathmoveto{\pgfqpoint{0.750000in}{2.602109in}}%
\pgfpathlineto{\pgfqpoint{5.413889in}{2.602109in}}%
\pgfusepath{stroke}%
\end{pgfscope}%
\begin{pgfscope}%
\pgfsetrectcap%
\pgfsetmiterjoin%
\pgfsetlinewidth{0.803000pt}%
\definecolor{currentstroke}{rgb}{0.000000,0.000000,0.000000}%
\pgfsetstrokecolor{currentstroke}%
\pgfsetdash{}{0pt}%
\pgfpathmoveto{\pgfqpoint{0.750000in}{0.500000in}}%
\pgfpathlineto{\pgfqpoint{0.750000in}{3.520000in}}%
\pgfusepath{stroke}%
\end{pgfscope}%
\begin{pgfscope}%
\pgfsetrectcap%
\pgfsetmiterjoin%
\pgfsetlinewidth{0.803000pt}%
\definecolor{currentstroke}{rgb}{0.000000,0.000000,0.000000}%
\pgfsetstrokecolor{currentstroke}%
\pgfsetdash{}{0pt}%
\pgfpathmoveto{\pgfqpoint{5.400000in}{0.500000in}}%
\pgfpathlineto{\pgfqpoint{5.400000in}{3.520000in}}%
\pgfusepath{stroke}%
\end{pgfscope}%
\begin{pgfscope}%
\pgfsetrectcap%
\pgfsetmiterjoin%
\pgfsetlinewidth{0.803000pt}%
\definecolor{currentstroke}{rgb}{0.000000,0.000000,0.000000}%
\pgfsetstrokecolor{currentstroke}%
\pgfsetdash{}{0pt}%
\pgfpathmoveto{\pgfqpoint{0.750000in}{0.500000in}}%
\pgfpathlineto{\pgfqpoint{5.400000in}{0.500000in}}%
\pgfusepath{stroke}%
\end{pgfscope}%
\begin{pgfscope}%
\pgfsetrectcap%
\pgfsetmiterjoin%
\pgfsetlinewidth{0.803000pt}%
\definecolor{currentstroke}{rgb}{0.000000,0.000000,0.000000}%
\pgfsetstrokecolor{currentstroke}%
\pgfsetdash{}{0pt}%
\pgfpathmoveto{\pgfqpoint{0.750000in}{3.520000in}}%
\pgfpathlineto{\pgfqpoint{5.400000in}{3.520000in}}%
\pgfusepath{stroke}%
\end{pgfscope}%
\begin{pgfscope}%
\definecolor{textcolor}{rgb}{0.000000,0.000000,0.000000}%
\pgfsetstrokecolor{textcolor}%
\pgfsetfillcolor{textcolor}%
\pgftext[x=3.075000in,y=3.603333in,,base]{\color{textcolor}\sffamily\fontsize{12.000000}{14.400000}\selectfont CIFAR10 VGG16 (Cross-entropy loss) }%
\end{pgfscope}%
\begin{pgfscope}%
\pgfsetbuttcap%
\pgfsetmiterjoin%
\definecolor{currentfill}{rgb}{1.000000,1.000000,1.000000}%
\pgfsetfillcolor{currentfill}%
\pgfsetfillopacity{0.800000}%
\pgfsetlinewidth{1.003750pt}%
\definecolor{currentstroke}{rgb}{0.800000,0.800000,0.800000}%
\pgfsetstrokecolor{currentstroke}%
\pgfsetstrokeopacity{0.800000}%
\pgfsetdash{}{0pt}%
\pgfpathmoveto{\pgfqpoint{3.291178in}{2.459815in}}%
\pgfpathlineto{\pgfqpoint{5.302778in}{2.459815in}}%
\pgfpathquadraticcurveto{\pgfqpoint{5.330556in}{2.459815in}}{\pgfqpoint{5.330556in}{2.487593in}}%
\pgfpathlineto{\pgfqpoint{5.330556in}{3.422778in}}%
\pgfpathquadraticcurveto{\pgfqpoint{5.330556in}{3.450556in}}{\pgfqpoint{5.302778in}{3.450556in}}%
\pgfpathlineto{\pgfqpoint{3.291178in}{3.450556in}}%
\pgfpathquadraticcurveto{\pgfqpoint{3.263400in}{3.450556in}}{\pgfqpoint{3.263400in}{3.422778in}}%
\pgfpathlineto{\pgfqpoint{3.263400in}{2.487593in}}%
\pgfpathquadraticcurveto{\pgfqpoint{3.263400in}{2.459815in}}{\pgfqpoint{3.291178in}{2.459815in}}%
\pgfpathclose%
\pgfusepath{stroke,fill}%
\end{pgfscope}%
\begin{pgfscope}%
\pgfsetrectcap%
\pgfsetroundjoin%
\pgfsetlinewidth{1.505625pt}%
\definecolor{currentstroke}{rgb}{0.121569,0.466667,0.705882}%
\pgfsetstrokecolor{currentstroke}%
\pgfsetdash{}{0pt}%
\pgfpathmoveto{\pgfqpoint{3.318956in}{3.346389in}}%
\pgfpathlineto{\pgfqpoint{3.596733in}{3.346389in}}%
\pgfusepath{stroke}%
\end{pgfscope}%
\begin{pgfscope}%
\definecolor{textcolor}{rgb}{0.000000,0.000000,0.000000}%
\pgfsetstrokecolor{textcolor}%
\pgfsetfillcolor{textcolor}%
\pgftext[x=3.707845in,y=3.297778in,left,base]{\color{textcolor}\sffamily\fontsize{10.000000}{12.000000}\selectfont Self learning run\# 1}%
\end{pgfscope}%
\begin{pgfscope}%
\pgfsetrectcap%
\pgfsetroundjoin%
\pgfsetlinewidth{1.505625pt}%
\definecolor{currentstroke}{rgb}{1.000000,0.498039,0.054902}%
\pgfsetstrokecolor{currentstroke}%
\pgfsetdash{}{0pt}%
\pgfpathmoveto{\pgfqpoint{3.318956in}{3.156574in}}%
\pgfpathlineto{\pgfqpoint{3.596733in}{3.156574in}}%
\pgfusepath{stroke}%
\end{pgfscope}%
\begin{pgfscope}%
\definecolor{textcolor}{rgb}{0.000000,0.000000,0.000000}%
\pgfsetstrokecolor{textcolor}%
\pgfsetfillcolor{textcolor}%
\pgftext[x=3.707845in,y=3.107963in,left,base]{\color{textcolor}\sffamily\fontsize{10.000000}{12.000000}\selectfont Self learning run\# 2}%
\end{pgfscope}%
\begin{pgfscope}%
\pgfsetrectcap%
\pgfsetroundjoin%
\pgfsetlinewidth{1.505625pt}%
\definecolor{currentstroke}{rgb}{0.172549,0.627451,0.172549}%
\pgfsetstrokecolor{currentstroke}%
\pgfsetdash{}{0pt}%
\pgfpathmoveto{\pgfqpoint{3.318956in}{2.966759in}}%
\pgfpathlineto{\pgfqpoint{3.596733in}{2.966759in}}%
\pgfusepath{stroke}%
\end{pgfscope}%
\begin{pgfscope}%
\definecolor{textcolor}{rgb}{0.000000,0.000000,0.000000}%
\pgfsetstrokecolor{textcolor}%
\pgfsetfillcolor{textcolor}%
\pgftext[x=3.707845in,y=2.918148in,left,base]{\color{textcolor}\sffamily\fontsize{10.000000}{12.000000}\selectfont Self learning run\# 3}%
\end{pgfscope}%
\begin{pgfscope}%
\pgfsetbuttcap%
\pgfsetroundjoin%
\pgfsetlinewidth{1.505625pt}%
\definecolor{currentstroke}{rgb}{0.000000,0.000000,1.000000}%
\pgfsetstrokecolor{currentstroke}%
\pgfsetdash{{5.550000pt}{2.400000pt}}{0.000000pt}%
\pgfpathmoveto{\pgfqpoint{3.318956in}{2.776945in}}%
\pgfpathlineto{\pgfqpoint{3.596733in}{2.776945in}}%
\pgfusepath{stroke}%
\end{pgfscope}%
\begin{pgfscope}%
\definecolor{textcolor}{rgb}{0.000000,0.000000,0.000000}%
\pgfsetstrokecolor{textcolor}%
\pgfsetfillcolor{textcolor}%
\pgftext[x=3.707845in,y=2.728334in,left,base]{\color{textcolor}\sffamily\fontsize{10.000000}{12.000000}\selectfont 4000-label 70.13 \(\displaystyle \pm\) 1.45\%}%
\end{pgfscope}%
\begin{pgfscope}%
\pgfsetbuttcap%
\pgfsetroundjoin%
\pgfsetlinewidth{1.505625pt}%
\definecolor{currentstroke}{rgb}{0.000000,0.500000,0.000000}%
\pgfsetstrokecolor{currentstroke}%
\pgfsetdash{{5.550000pt}{2.400000pt}}{0.000000pt}%
\pgfpathmoveto{\pgfqpoint{3.318956in}{2.587130in}}%
\pgfpathlineto{\pgfqpoint{3.596733in}{2.587130in}}%
\pgfusepath{stroke}%
\end{pgfscope}%
\begin{pgfscope}%
\definecolor{textcolor}{rgb}{0.000000,0.000000,0.000000}%
\pgfsetstrokecolor{textcolor}%
\pgfsetfillcolor{textcolor}%
\pgftext[x=3.707845in,y=2.538519in,left,base]{\color{textcolor}\sffamily\fontsize{10.000000}{12.000000}\selectfont All label 87.84 \(\displaystyle \pm\) 0.39\%}%
\end{pgfscope}%
\end{pgfpicture}%
\makeatother%
\endgroup%

%% file: imgs/pretrained_weights.pgf
\begingroup%
\makeatletter%
\begin{pgfpicture}%
\pgfpathrectangle{\pgfpointorigin}{\pgfqpoint{6.000000in}{4.000000in}}%
\pgfusepath{use as bounding box, clip}%
\begin{pgfscope}%
\pgfsetbuttcap%
\pgfsetmiterjoin%
\definecolor{currentfill}{rgb}{1.000000,1.000000,1.000000}%
\pgfsetfillcolor{currentfill}%
\pgfsetlinewidth{0.000000pt}%
\definecolor{currentstroke}{rgb}{1.000000,1.000000,1.000000}%
\pgfsetstrokecolor{currentstroke}%
\pgfsetdash{}{0pt}%
\pgfpathmoveto{\pgfqpoint{0.000000in}{0.000000in}}%
\pgfpathlineto{\pgfqpoint{6.000000in}{0.000000in}}%
\pgfpathlineto{\pgfqpoint{6.000000in}{4.000000in}}%
\pgfpathlineto{\pgfqpoint{0.000000in}{4.000000in}}%
\pgfpathclose%
\pgfusepath{fill}%
\end{pgfscope}%
\begin{pgfscope}%
\pgfsetbuttcap%
\pgfsetmiterjoin%
\definecolor{currentfill}{rgb}{1.000000,1.000000,1.000000}%
\pgfsetfillcolor{currentfill}%
\pgfsetlinewidth{0.000000pt}%
\definecolor{currentstroke}{rgb}{0.000000,0.000000,0.000000}%
\pgfsetstrokecolor{currentstroke}%
\pgfsetstrokeopacity{0.000000}%
\pgfsetdash{}{0pt}%
\pgfpathmoveto{\pgfqpoint{0.750000in}{0.500000in}}%
\pgfpathlineto{\pgfqpoint{5.400000in}{0.500000in}}%
\pgfpathlineto{\pgfqpoint{5.400000in}{3.520000in}}%
\pgfpathlineto{\pgfqpoint{0.750000in}{3.520000in}}%
\pgfpathclose%
\pgfusepath{fill}%
\end{pgfscope}%
\begin{pgfscope}%
\pgfpathrectangle{\pgfqpoint{0.750000in}{0.500000in}}{\pgfqpoint{4.650000in}{3.020000in}}%
\pgfusepath{clip}%
\pgfsetrectcap%
\pgfsetroundjoin%
\pgfsetlinewidth{0.803000pt}%
\definecolor{currentstroke}{rgb}{0.690196,0.690196,0.690196}%
\pgfsetstrokecolor{currentstroke}%
\pgfsetdash{}{0pt}%
\pgfpathmoveto{\pgfqpoint{0.961364in}{0.500000in}}%
\pgfpathlineto{\pgfqpoint{0.961364in}{3.520000in}}%
\pgfusepath{stroke}%
\end{pgfscope}%
\begin{pgfscope}%
\pgfsetbuttcap%
\pgfsetroundjoin%
\definecolor{currentfill}{rgb}{0.000000,0.000000,0.000000}%
\pgfsetfillcolor{currentfill}%
\pgfsetlinewidth{0.803000pt}%
\definecolor{currentstroke}{rgb}{0.000000,0.000000,0.000000}%
\pgfsetstrokecolor{currentstroke}%
\pgfsetdash{}{0pt}%
\pgfsys@defobject{currentmarker}{\pgfqpoint{0.000000in}{-0.048611in}}{\pgfqpoint{0.000000in}{0.000000in}}{%
\pgfpathmoveto{\pgfqpoint{0.000000in}{0.000000in}}%
\pgfpathlineto{\pgfqpoint{0.000000in}{-0.048611in}}%
\pgfusepath{stroke,fill}%
}%
\begin{pgfscope}%
\pgfsys@transformshift{0.961364in}{0.500000in}%
\pgfsys@useobject{currentmarker}{}%
\end{pgfscope}%
\end{pgfscope}%
\begin{pgfscope}%
\definecolor{textcolor}{rgb}{0.000000,0.000000,0.000000}%
\pgfsetstrokecolor{textcolor}%
\pgfsetfillcolor{textcolor}%
\pgftext[x=0.961364in,y=0.402778in,,top]{\color{textcolor}\sffamily\fontsize{10.000000}{12.000000}\selectfont \(\displaystyle {0}\)}%
\end{pgfscope}%
\begin{pgfscope}%
\pgfpathrectangle{\pgfqpoint{0.750000in}{0.500000in}}{\pgfqpoint{4.650000in}{3.020000in}}%
\pgfusepath{clip}%
\pgfsetrectcap%
\pgfsetroundjoin%
\pgfsetlinewidth{0.803000pt}%
\definecolor{currentstroke}{rgb}{0.690196,0.690196,0.690196}%
\pgfsetstrokecolor{currentstroke}%
\pgfsetdash{}{0pt}%
\pgfpathmoveto{\pgfqpoint{1.842045in}{0.500000in}}%
\pgfpathlineto{\pgfqpoint{1.842045in}{3.520000in}}%
\pgfusepath{stroke}%
\end{pgfscope}%
\begin{pgfscope}%
\pgfsetbuttcap%
\pgfsetroundjoin%
\definecolor{currentfill}{rgb}{0.000000,0.000000,0.000000}%
\pgfsetfillcolor{currentfill}%
\pgfsetlinewidth{0.803000pt}%
\definecolor{currentstroke}{rgb}{0.000000,0.000000,0.000000}%
\pgfsetstrokecolor{currentstroke}%
\pgfsetdash{}{0pt}%
\pgfsys@defobject{currentmarker}{\pgfqpoint{0.000000in}{-0.048611in}}{\pgfqpoint{0.000000in}{0.000000in}}{%
\pgfpathmoveto{\pgfqpoint{0.000000in}{0.000000in}}%
\pgfpathlineto{\pgfqpoint{0.000000in}{-0.048611in}}%
\pgfusepath{stroke,fill}%
}%
\begin{pgfscope}%
\pgfsys@transformshift{1.842045in}{0.500000in}%
\pgfsys@useobject{currentmarker}{}%
\end{pgfscope}%
\end{pgfscope}%
\begin{pgfscope}%
\definecolor{textcolor}{rgb}{0.000000,0.000000,0.000000}%
\pgfsetstrokecolor{textcolor}%
\pgfsetfillcolor{textcolor}%
\pgftext[x=1.842045in,y=0.402778in,,top]{\color{textcolor}\sffamily\fontsize{10.000000}{12.000000}\selectfont \(\displaystyle {5}\)}%
\end{pgfscope}%
\begin{pgfscope}%
\pgfpathrectangle{\pgfqpoint{0.750000in}{0.500000in}}{\pgfqpoint{4.650000in}{3.020000in}}%
\pgfusepath{clip}%
\pgfsetrectcap%
\pgfsetroundjoin%
\pgfsetlinewidth{0.803000pt}%
\definecolor{currentstroke}{rgb}{0.690196,0.690196,0.690196}%
\pgfsetstrokecolor{currentstroke}%
\pgfsetdash{}{0pt}%
\pgfpathmoveto{\pgfqpoint{2.722727in}{0.500000in}}%
\pgfpathlineto{\pgfqpoint{2.722727in}{3.520000in}}%
\pgfusepath{stroke}%
\end{pgfscope}%
\begin{pgfscope}%
\pgfsetbuttcap%
\pgfsetroundjoin%
\definecolor{currentfill}{rgb}{0.000000,0.000000,0.000000}%
\pgfsetfillcolor{currentfill}%
\pgfsetlinewidth{0.803000pt}%
\definecolor{currentstroke}{rgb}{0.000000,0.000000,0.000000}%
\pgfsetstrokecolor{currentstroke}%
\pgfsetdash{}{0pt}%
\pgfsys@defobject{currentmarker}{\pgfqpoint{0.000000in}{-0.048611in}}{\pgfqpoint{0.000000in}{0.000000in}}{%
\pgfpathmoveto{\pgfqpoint{0.000000in}{0.000000in}}%
\pgfpathlineto{\pgfqpoint{0.000000in}{-0.048611in}}%
\pgfusepath{stroke,fill}%
}%
\begin{pgfscope}%
\pgfsys@transformshift{2.722727in}{0.500000in}%
\pgfsys@useobject{currentmarker}{}%
\end{pgfscope}%
\end{pgfscope}%
\begin{pgfscope}%
\definecolor{textcolor}{rgb}{0.000000,0.000000,0.000000}%
\pgfsetstrokecolor{textcolor}%
\pgfsetfillcolor{textcolor}%
\pgftext[x=2.722727in,y=0.402778in,,top]{\color{textcolor}\sffamily\fontsize{10.000000}{12.000000}\selectfont \(\displaystyle {10}\)}%
\end{pgfscope}%
\begin{pgfscope}%
\pgfpathrectangle{\pgfqpoint{0.750000in}{0.500000in}}{\pgfqpoint{4.650000in}{3.020000in}}%
\pgfusepath{clip}%
\pgfsetrectcap%
\pgfsetroundjoin%
\pgfsetlinewidth{0.803000pt}%
\definecolor{currentstroke}{rgb}{0.690196,0.690196,0.690196}%
\pgfsetstrokecolor{currentstroke}%
\pgfsetdash{}{0pt}%
\pgfpathmoveto{\pgfqpoint{3.603409in}{0.500000in}}%
\pgfpathlineto{\pgfqpoint{3.603409in}{3.520000in}}%
\pgfusepath{stroke}%
\end{pgfscope}%
\begin{pgfscope}%
\pgfsetbuttcap%
\pgfsetroundjoin%
\definecolor{currentfill}{rgb}{0.000000,0.000000,0.000000}%
\pgfsetfillcolor{currentfill}%
\pgfsetlinewidth{0.803000pt}%
\definecolor{currentstroke}{rgb}{0.000000,0.000000,0.000000}%
\pgfsetstrokecolor{currentstroke}%
\pgfsetdash{}{0pt}%
\pgfsys@defobject{currentmarker}{\pgfqpoint{0.000000in}{-0.048611in}}{\pgfqpoint{0.000000in}{0.000000in}}{%
\pgfpathmoveto{\pgfqpoint{0.000000in}{0.000000in}}%
\pgfpathlineto{\pgfqpoint{0.000000in}{-0.048611in}}%
\pgfusepath{stroke,fill}%
}%
\begin{pgfscope}%
\pgfsys@transformshift{3.603409in}{0.500000in}%
\pgfsys@useobject{currentmarker}{}%
\end{pgfscope}%
\end{pgfscope}%
\begin{pgfscope}%
\definecolor{textcolor}{rgb}{0.000000,0.000000,0.000000}%
\pgfsetstrokecolor{textcolor}%
\pgfsetfillcolor{textcolor}%
\pgftext[x=3.603409in,y=0.402778in,,top]{\color{textcolor}\sffamily\fontsize{10.000000}{12.000000}\selectfont \(\displaystyle {15}\)}%
\end{pgfscope}%
\begin{pgfscope}%
\pgfpathrectangle{\pgfqpoint{0.750000in}{0.500000in}}{\pgfqpoint{4.650000in}{3.020000in}}%
\pgfusepath{clip}%
\pgfsetrectcap%
\pgfsetroundjoin%
\pgfsetlinewidth{0.803000pt}%
\definecolor{currentstroke}{rgb}{0.690196,0.690196,0.690196}%
\pgfsetstrokecolor{currentstroke}%
\pgfsetdash{}{0pt}%
\pgfpathmoveto{\pgfqpoint{4.484091in}{0.500000in}}%
\pgfpathlineto{\pgfqpoint{4.484091in}{3.520000in}}%
\pgfusepath{stroke}%
\end{pgfscope}%
\begin{pgfscope}%
\pgfsetbuttcap%
\pgfsetroundjoin%
\definecolor{currentfill}{rgb}{0.000000,0.000000,0.000000}%
\pgfsetfillcolor{currentfill}%
\pgfsetlinewidth{0.803000pt}%
\definecolor{currentstroke}{rgb}{0.000000,0.000000,0.000000}%
\pgfsetstrokecolor{currentstroke}%
\pgfsetdash{}{0pt}%
\pgfsys@defobject{currentmarker}{\pgfqpoint{0.000000in}{-0.048611in}}{\pgfqpoint{0.000000in}{0.000000in}}{%
\pgfpathmoveto{\pgfqpoint{0.000000in}{0.000000in}}%
\pgfpathlineto{\pgfqpoint{0.000000in}{-0.048611in}}%
\pgfusepath{stroke,fill}%
}%
\begin{pgfscope}%
\pgfsys@transformshift{4.484091in}{0.500000in}%
\pgfsys@useobject{currentmarker}{}%
\end{pgfscope}%
\end{pgfscope}%
\begin{pgfscope}%
\definecolor{textcolor}{rgb}{0.000000,0.000000,0.000000}%
\pgfsetstrokecolor{textcolor}%
\pgfsetfillcolor{textcolor}%
\pgftext[x=4.484091in,y=0.402778in,,top]{\color{textcolor}\sffamily\fontsize{10.000000}{12.000000}\selectfont \(\displaystyle {20}\)}%
\end{pgfscope}%
\begin{pgfscope}%
\pgfpathrectangle{\pgfqpoint{0.750000in}{0.500000in}}{\pgfqpoint{4.650000in}{3.020000in}}%
\pgfusepath{clip}%
\pgfsetrectcap%
\pgfsetroundjoin%
\pgfsetlinewidth{0.803000pt}%
\definecolor{currentstroke}{rgb}{0.690196,0.690196,0.690196}%
\pgfsetstrokecolor{currentstroke}%
\pgfsetdash{}{0pt}%
\pgfpathmoveto{\pgfqpoint{5.364773in}{0.500000in}}%
\pgfpathlineto{\pgfqpoint{5.364773in}{3.520000in}}%
\pgfusepath{stroke}%
\end{pgfscope}%
\begin{pgfscope}%
\pgfsetbuttcap%
\pgfsetroundjoin%
\definecolor{currentfill}{rgb}{0.000000,0.000000,0.000000}%
\pgfsetfillcolor{currentfill}%
\pgfsetlinewidth{0.803000pt}%
\definecolor{currentstroke}{rgb}{0.000000,0.000000,0.000000}%
\pgfsetstrokecolor{currentstroke}%
\pgfsetdash{}{0pt}%
\pgfsys@defobject{currentmarker}{\pgfqpoint{0.000000in}{-0.048611in}}{\pgfqpoint{0.000000in}{0.000000in}}{%
\pgfpathmoveto{\pgfqpoint{0.000000in}{0.000000in}}%
\pgfpathlineto{\pgfqpoint{0.000000in}{-0.048611in}}%
\pgfusepath{stroke,fill}%
}%
\begin{pgfscope}%
\pgfsys@transformshift{5.364773in}{0.500000in}%
\pgfsys@useobject{currentmarker}{}%
\end{pgfscope}%
\end{pgfscope}%
\begin{pgfscope}%
\definecolor{textcolor}{rgb}{0.000000,0.000000,0.000000}%
\pgfsetstrokecolor{textcolor}%
\pgfsetfillcolor{textcolor}%
\pgftext[x=5.364773in,y=0.402778in,,top]{\color{textcolor}\sffamily\fontsize{10.000000}{12.000000}\selectfont \(\displaystyle {25}\)}%
\end{pgfscope}%
\begin{pgfscope}%
\definecolor{textcolor}{rgb}{0.000000,0.000000,0.000000}%
\pgfsetstrokecolor{textcolor}%
\pgfsetfillcolor{textcolor}%
\pgftext[x=3.075000in,y=0.227624in,,top]{\color{textcolor}\sffamily\fontsize{10.000000}{12.000000}\selectfont Meta Iterations}%
\end{pgfscope}%
\begin{pgfscope}%
\pgfpathrectangle{\pgfqpoint{0.750000in}{0.500000in}}{\pgfqpoint{4.650000in}{3.020000in}}%
\pgfusepath{clip}%
\pgfsetrectcap%
\pgfsetroundjoin%
\pgfsetlinewidth{0.803000pt}%
\definecolor{currentstroke}{rgb}{0.690196,0.690196,0.690196}%
\pgfsetstrokecolor{currentstroke}%
\pgfsetdash{}{0pt}%
\pgfpathmoveto{\pgfqpoint{0.750000in}{0.500000in}}%
\pgfpathlineto{\pgfqpoint{5.400000in}{0.500000in}}%
\pgfusepath{stroke}%
\end{pgfscope}%
\begin{pgfscope}%
\pgfsetbuttcap%
\pgfsetroundjoin%
\definecolor{currentfill}{rgb}{0.000000,0.000000,0.000000}%
\pgfsetfillcolor{currentfill}%
\pgfsetlinewidth{0.803000pt}%
\definecolor{currentstroke}{rgb}{0.000000,0.000000,0.000000}%
\pgfsetstrokecolor{currentstroke}%
\pgfsetdash{}{0pt}%
\pgfsys@defobject{currentmarker}{\pgfqpoint{-0.048611in}{0.000000in}}{\pgfqpoint{0.000000in}{0.000000in}}{%
\pgfpathmoveto{\pgfqpoint{0.000000in}{0.000000in}}%
\pgfpathlineto{\pgfqpoint{-0.048611in}{0.000000in}}%
\pgfusepath{stroke,fill}%
}%
\begin{pgfscope}%
\pgfsys@transformshift{0.750000in}{0.500000in}%
\pgfsys@useobject{currentmarker}{}%
\end{pgfscope}%
\end{pgfscope}%
\begin{pgfscope}%
\definecolor{textcolor}{rgb}{0.000000,0.000000,0.000000}%
\pgfsetstrokecolor{textcolor}%
\pgfsetfillcolor{textcolor}%
\pgftext[x=0.506946in, y=0.451775in, left, base]{\color{textcolor}\sffamily\fontsize{10.000000}{12.000000}\selectfont \(\displaystyle {60}\)}%
\end{pgfscope}%
\begin{pgfscope}%
\pgfpathrectangle{\pgfqpoint{0.750000in}{0.500000in}}{\pgfqpoint{4.650000in}{3.020000in}}%
\pgfusepath{clip}%
\pgfsetrectcap%
\pgfsetroundjoin%
\pgfsetlinewidth{0.803000pt}%
\definecolor{currentstroke}{rgb}{0.690196,0.690196,0.690196}%
\pgfsetstrokecolor{currentstroke}%
\pgfsetdash{}{0pt}%
\pgfpathmoveto{\pgfqpoint{0.750000in}{0.877500in}}%
\pgfpathlineto{\pgfqpoint{5.400000in}{0.877500in}}%
\pgfusepath{stroke}%
\end{pgfscope}%
\begin{pgfscope}%
\pgfsetbuttcap%
\pgfsetroundjoin%
\definecolor{currentfill}{rgb}{0.000000,0.000000,0.000000}%
\pgfsetfillcolor{currentfill}%
\pgfsetlinewidth{0.803000pt}%
\definecolor{currentstroke}{rgb}{0.000000,0.000000,0.000000}%
\pgfsetstrokecolor{currentstroke}%
\pgfsetdash{}{0pt}%
\pgfsys@defobject{currentmarker}{\pgfqpoint{-0.048611in}{0.000000in}}{\pgfqpoint{0.000000in}{0.000000in}}{%
\pgfpathmoveto{\pgfqpoint{0.000000in}{0.000000in}}%
\pgfpathlineto{\pgfqpoint{-0.048611in}{0.000000in}}%
\pgfusepath{stroke,fill}%
}%
\begin{pgfscope}%
\pgfsys@transformshift{0.750000in}{0.877500in}%
\pgfsys@useobject{currentmarker}{}%
\end{pgfscope}%
\end{pgfscope}%
\begin{pgfscope}%
\definecolor{textcolor}{rgb}{0.000000,0.000000,0.000000}%
\pgfsetstrokecolor{textcolor}%
\pgfsetfillcolor{textcolor}%
\pgftext[x=0.506946in, y=0.829275in, left, base]{\color{textcolor}\sffamily\fontsize{10.000000}{12.000000}\selectfont \(\displaystyle {65}\)}%
\end{pgfscope}%
\begin{pgfscope}%
\pgfpathrectangle{\pgfqpoint{0.750000in}{0.500000in}}{\pgfqpoint{4.650000in}{3.020000in}}%
\pgfusepath{clip}%
\pgfsetrectcap%
\pgfsetroundjoin%
\pgfsetlinewidth{0.803000pt}%
\definecolor{currentstroke}{rgb}{0.690196,0.690196,0.690196}%
\pgfsetstrokecolor{currentstroke}%
\pgfsetdash{}{0pt}%
\pgfpathmoveto{\pgfqpoint{0.750000in}{1.255000in}}%
\pgfpathlineto{\pgfqpoint{5.400000in}{1.255000in}}%
\pgfusepath{stroke}%
\end{pgfscope}%
\begin{pgfscope}%
\pgfsetbuttcap%
\pgfsetroundjoin%
\definecolor{currentfill}{rgb}{0.000000,0.000000,0.000000}%
\pgfsetfillcolor{currentfill}%
\pgfsetlinewidth{0.803000pt}%
\definecolor{currentstroke}{rgb}{0.000000,0.000000,0.000000}%
\pgfsetstrokecolor{currentstroke}%
\pgfsetdash{}{0pt}%
\pgfsys@defobject{currentmarker}{\pgfqpoint{-0.048611in}{0.000000in}}{\pgfqpoint{0.000000in}{0.000000in}}{%
\pgfpathmoveto{\pgfqpoint{0.000000in}{0.000000in}}%
\pgfpathlineto{\pgfqpoint{-0.048611in}{0.000000in}}%
\pgfusepath{stroke,fill}%
}%
\begin{pgfscope}%
\pgfsys@transformshift{0.750000in}{1.255000in}%
\pgfsys@useobject{currentmarker}{}%
\end{pgfscope}%
\end{pgfscope}%
\begin{pgfscope}%
\definecolor{textcolor}{rgb}{0.000000,0.000000,0.000000}%
\pgfsetstrokecolor{textcolor}%
\pgfsetfillcolor{textcolor}%
\pgftext[x=0.506946in, y=1.206775in, left, base]{\color{textcolor}\sffamily\fontsize{10.000000}{12.000000}\selectfont \(\displaystyle {70}\)}%
\end{pgfscope}%
\begin{pgfscope}%
\pgfpathrectangle{\pgfqpoint{0.750000in}{0.500000in}}{\pgfqpoint{4.650000in}{3.020000in}}%
\pgfusepath{clip}%
\pgfsetrectcap%
\pgfsetroundjoin%
\pgfsetlinewidth{0.803000pt}%
\definecolor{currentstroke}{rgb}{0.690196,0.690196,0.690196}%
\pgfsetstrokecolor{currentstroke}%
\pgfsetdash{}{0pt}%
\pgfpathmoveto{\pgfqpoint{0.750000in}{1.632500in}}%
\pgfpathlineto{\pgfqpoint{5.400000in}{1.632500in}}%
\pgfusepath{stroke}%
\end{pgfscope}%
\begin{pgfscope}%
\pgfsetbuttcap%
\pgfsetroundjoin%
\definecolor{currentfill}{rgb}{0.000000,0.000000,0.000000}%
\pgfsetfillcolor{currentfill}%
\pgfsetlinewidth{0.803000pt}%
\definecolor{currentstroke}{rgb}{0.000000,0.000000,0.000000}%
\pgfsetstrokecolor{currentstroke}%
\pgfsetdash{}{0pt}%
\pgfsys@defobject{currentmarker}{\pgfqpoint{-0.048611in}{0.000000in}}{\pgfqpoint{0.000000in}{0.000000in}}{%
\pgfpathmoveto{\pgfqpoint{0.000000in}{0.000000in}}%
\pgfpathlineto{\pgfqpoint{-0.048611in}{0.000000in}}%
\pgfusepath{stroke,fill}%
}%
\begin{pgfscope}%
\pgfsys@transformshift{0.750000in}{1.632500in}%
\pgfsys@useobject{currentmarker}{}%
\end{pgfscope}%
\end{pgfscope}%
\begin{pgfscope}%
\definecolor{textcolor}{rgb}{0.000000,0.000000,0.000000}%
\pgfsetstrokecolor{textcolor}%
\pgfsetfillcolor{textcolor}%
\pgftext[x=0.506946in, y=1.584275in, left, base]{\color{textcolor}\sffamily\fontsize{10.000000}{12.000000}\selectfont \(\displaystyle {75}\)}%
\end{pgfscope}%
\begin{pgfscope}%
\pgfpathrectangle{\pgfqpoint{0.750000in}{0.500000in}}{\pgfqpoint{4.650000in}{3.020000in}}%
\pgfusepath{clip}%
\pgfsetrectcap%
\pgfsetroundjoin%
\pgfsetlinewidth{0.803000pt}%
\definecolor{currentstroke}{rgb}{0.690196,0.690196,0.690196}%
\pgfsetstrokecolor{currentstroke}%
\pgfsetdash{}{0pt}%
\pgfpathmoveto{\pgfqpoint{0.750000in}{2.010000in}}%
\pgfpathlineto{\pgfqpoint{5.400000in}{2.010000in}}%
\pgfusepath{stroke}%
\end{pgfscope}%
\begin{pgfscope}%
\pgfsetbuttcap%
\pgfsetroundjoin%
\definecolor{currentfill}{rgb}{0.000000,0.000000,0.000000}%
\pgfsetfillcolor{currentfill}%
\pgfsetlinewidth{0.803000pt}%
\definecolor{currentstroke}{rgb}{0.000000,0.000000,0.000000}%
\pgfsetstrokecolor{currentstroke}%
\pgfsetdash{}{0pt}%
\pgfsys@defobject{currentmarker}{\pgfqpoint{-0.048611in}{0.000000in}}{\pgfqpoint{0.000000in}{0.000000in}}{%
\pgfpathmoveto{\pgfqpoint{0.000000in}{0.000000in}}%
\pgfpathlineto{\pgfqpoint{-0.048611in}{0.000000in}}%
\pgfusepath{stroke,fill}%
}%
\begin{pgfscope}%
\pgfsys@transformshift{0.750000in}{2.010000in}%
\pgfsys@useobject{currentmarker}{}%
\end{pgfscope}%
\end{pgfscope}%
\begin{pgfscope}%
\definecolor{textcolor}{rgb}{0.000000,0.000000,0.000000}%
\pgfsetstrokecolor{textcolor}%
\pgfsetfillcolor{textcolor}%
\pgftext[x=0.506946in, y=1.961775in, left, base]{\color{textcolor}\sffamily\fontsize{10.000000}{12.000000}\selectfont \(\displaystyle {80}\)}%
\end{pgfscope}%
\begin{pgfscope}%
\pgfpathrectangle{\pgfqpoint{0.750000in}{0.500000in}}{\pgfqpoint{4.650000in}{3.020000in}}%
\pgfusepath{clip}%
\pgfsetrectcap%
\pgfsetroundjoin%
\pgfsetlinewidth{0.803000pt}%
\definecolor{currentstroke}{rgb}{0.690196,0.690196,0.690196}%
\pgfsetstrokecolor{currentstroke}%
\pgfsetdash{}{0pt}%
\pgfpathmoveto{\pgfqpoint{0.750000in}{2.387500in}}%
\pgfpathlineto{\pgfqpoint{5.400000in}{2.387500in}}%
\pgfusepath{stroke}%
\end{pgfscope}%
\begin{pgfscope}%
\pgfsetbuttcap%
\pgfsetroundjoin%
\definecolor{currentfill}{rgb}{0.000000,0.000000,0.000000}%
\pgfsetfillcolor{currentfill}%
\pgfsetlinewidth{0.803000pt}%
\definecolor{currentstroke}{rgb}{0.000000,0.000000,0.000000}%
\pgfsetstrokecolor{currentstroke}%
\pgfsetdash{}{0pt}%
\pgfsys@defobject{currentmarker}{\pgfqpoint{-0.048611in}{0.000000in}}{\pgfqpoint{0.000000in}{0.000000in}}{%
\pgfpathmoveto{\pgfqpoint{0.000000in}{0.000000in}}%
\pgfpathlineto{\pgfqpoint{-0.048611in}{0.000000in}}%
\pgfusepath{stroke,fill}%
}%
\begin{pgfscope}%
\pgfsys@transformshift{0.750000in}{2.387500in}%
\pgfsys@useobject{currentmarker}{}%
\end{pgfscope}%
\end{pgfscope}%
\begin{pgfscope}%
\definecolor{textcolor}{rgb}{0.000000,0.000000,0.000000}%
\pgfsetstrokecolor{textcolor}%
\pgfsetfillcolor{textcolor}%
\pgftext[x=0.506946in, y=2.339275in, left, base]{\color{textcolor}\sffamily\fontsize{10.000000}{12.000000}\selectfont \(\displaystyle {85}\)}%
\end{pgfscope}%
\begin{pgfscope}%
\pgfpathrectangle{\pgfqpoint{0.750000in}{0.500000in}}{\pgfqpoint{4.650000in}{3.020000in}}%
\pgfusepath{clip}%
\pgfsetrectcap%
\pgfsetroundjoin%
\pgfsetlinewidth{0.803000pt}%
\definecolor{currentstroke}{rgb}{0.690196,0.690196,0.690196}%
\pgfsetstrokecolor{currentstroke}%
\pgfsetdash{}{0pt}%
\pgfpathmoveto{\pgfqpoint{0.750000in}{2.765000in}}%
\pgfpathlineto{\pgfqpoint{5.400000in}{2.765000in}}%
\pgfusepath{stroke}%
\end{pgfscope}%
\begin{pgfscope}%
\pgfsetbuttcap%
\pgfsetroundjoin%
\definecolor{currentfill}{rgb}{0.000000,0.000000,0.000000}%
\pgfsetfillcolor{currentfill}%
\pgfsetlinewidth{0.803000pt}%
\definecolor{currentstroke}{rgb}{0.000000,0.000000,0.000000}%
\pgfsetstrokecolor{currentstroke}%
\pgfsetdash{}{0pt}%
\pgfsys@defobject{currentmarker}{\pgfqpoint{-0.048611in}{0.000000in}}{\pgfqpoint{0.000000in}{0.000000in}}{%
\pgfpathmoveto{\pgfqpoint{0.000000in}{0.000000in}}%
\pgfpathlineto{\pgfqpoint{-0.048611in}{0.000000in}}%
\pgfusepath{stroke,fill}%
}%
\begin{pgfscope}%
\pgfsys@transformshift{0.750000in}{2.765000in}%
\pgfsys@useobject{currentmarker}{}%
\end{pgfscope}%
\end{pgfscope}%
\begin{pgfscope}%
\definecolor{textcolor}{rgb}{0.000000,0.000000,0.000000}%
\pgfsetstrokecolor{textcolor}%
\pgfsetfillcolor{textcolor}%
\pgftext[x=0.506946in, y=2.716775in, left, base]{\color{textcolor}\sffamily\fontsize{10.000000}{12.000000}\selectfont \(\displaystyle {90}\)}%
\end{pgfscope}%
\begin{pgfscope}%
\pgfpathrectangle{\pgfqpoint{0.750000in}{0.500000in}}{\pgfqpoint{4.650000in}{3.020000in}}%
\pgfusepath{clip}%
\pgfsetrectcap%
\pgfsetroundjoin%
\pgfsetlinewidth{0.803000pt}%
\definecolor{currentstroke}{rgb}{0.690196,0.690196,0.690196}%
\pgfsetstrokecolor{currentstroke}%
\pgfsetdash{}{0pt}%
\pgfpathmoveto{\pgfqpoint{0.750000in}{3.142500in}}%
\pgfpathlineto{\pgfqpoint{5.400000in}{3.142500in}}%
\pgfusepath{stroke}%
\end{pgfscope}%
\begin{pgfscope}%
\pgfsetbuttcap%
\pgfsetroundjoin%
\definecolor{currentfill}{rgb}{0.000000,0.000000,0.000000}%
\pgfsetfillcolor{currentfill}%
\pgfsetlinewidth{0.803000pt}%
\definecolor{currentstroke}{rgb}{0.000000,0.000000,0.000000}%
\pgfsetstrokecolor{currentstroke}%
\pgfsetdash{}{0pt}%
\pgfsys@defobject{currentmarker}{\pgfqpoint{-0.048611in}{0.000000in}}{\pgfqpoint{0.000000in}{0.000000in}}{%
\pgfpathmoveto{\pgfqpoint{0.000000in}{0.000000in}}%
\pgfpathlineto{\pgfqpoint{-0.048611in}{0.000000in}}%
\pgfusepath{stroke,fill}%
}%
\begin{pgfscope}%
\pgfsys@transformshift{0.750000in}{3.142500in}%
\pgfsys@useobject{currentmarker}{}%
\end{pgfscope}%
\end{pgfscope}%
\begin{pgfscope}%
\definecolor{textcolor}{rgb}{0.000000,0.000000,0.000000}%
\pgfsetstrokecolor{textcolor}%
\pgfsetfillcolor{textcolor}%
\pgftext[x=0.506946in, y=3.094275in, left, base]{\color{textcolor}\sffamily\fontsize{10.000000}{12.000000}\selectfont \(\displaystyle {95}\)}%
\end{pgfscope}%
\begin{pgfscope}%
\pgfpathrectangle{\pgfqpoint{0.750000in}{0.500000in}}{\pgfqpoint{4.650000in}{3.020000in}}%
\pgfusepath{clip}%
\pgfsetrectcap%
\pgfsetroundjoin%
\pgfsetlinewidth{0.803000pt}%
\definecolor{currentstroke}{rgb}{0.690196,0.690196,0.690196}%
\pgfsetstrokecolor{currentstroke}%
\pgfsetdash{}{0pt}%
\pgfpathmoveto{\pgfqpoint{0.750000in}{3.520000in}}%
\pgfpathlineto{\pgfqpoint{5.400000in}{3.520000in}}%
\pgfusepath{stroke}%
\end{pgfscope}%
\begin{pgfscope}%
\pgfsetbuttcap%
\pgfsetroundjoin%
\definecolor{currentfill}{rgb}{0.000000,0.000000,0.000000}%
\pgfsetfillcolor{currentfill}%
\pgfsetlinewidth{0.803000pt}%
\definecolor{currentstroke}{rgb}{0.000000,0.000000,0.000000}%
\pgfsetstrokecolor{currentstroke}%
\pgfsetdash{}{0pt}%
\pgfsys@defobject{currentmarker}{\pgfqpoint{-0.048611in}{0.000000in}}{\pgfqpoint{0.000000in}{0.000000in}}{%
\pgfpathmoveto{\pgfqpoint{0.000000in}{0.000000in}}%
\pgfpathlineto{\pgfqpoint{-0.048611in}{0.000000in}}%
\pgfusepath{stroke,fill}%
}%
\begin{pgfscope}%
\pgfsys@transformshift{0.750000in}{3.520000in}%
\pgfsys@useobject{currentmarker}{}%
\end{pgfscope}%
\end{pgfscope}%
\begin{pgfscope}%
\definecolor{textcolor}{rgb}{0.000000,0.000000,0.000000}%
\pgfsetstrokecolor{textcolor}%
\pgfsetfillcolor{textcolor}%
\pgftext[x=0.434030in, y=3.471775in, left, base]{\color{textcolor}\sffamily\fontsize{10.000000}{12.000000}\selectfont \(\displaystyle {100}\)}%
\end{pgfscope}%
\begin{pgfscope}%
\definecolor{textcolor}{rgb}{0.000000,0.000000,0.000000}%
\pgfsetstrokecolor{textcolor}%
\pgfsetfillcolor{textcolor}%
\pgftext[x=0.378474in,y=2.010000in,,bottom,rotate=90.000000]{\color{textcolor}\sffamily\fontsize{10.000000}{12.000000}\selectfont Accuracy \%}%
\end{pgfscope}%
\begin{pgfscope}%
\pgfpathrectangle{\pgfqpoint{0.750000in}{0.500000in}}{\pgfqpoint{4.650000in}{3.020000in}}%
\pgfusepath{clip}%
\pgfsetrectcap%
\pgfsetroundjoin%
\pgfsetlinewidth{1.505625pt}%
\definecolor{currentstroke}{rgb}{0.121569,0.466667,0.705882}%
\pgfsetstrokecolor{currentstroke}%
\pgfsetdash{}{0pt}%
\pgfpathmoveto{\pgfqpoint{0.961364in}{1.780480in}}%
\pgfpathlineto{\pgfqpoint{1.137500in}{1.818230in}}%
\pgfpathlineto{\pgfqpoint{1.313636in}{1.963945in}}%
\pgfpathlineto{\pgfqpoint{1.489773in}{1.966210in}}%
\pgfpathlineto{\pgfqpoint{1.665909in}{2.022080in}}%
\pgfpathlineto{\pgfqpoint{1.842045in}{2.051525in}}%
\pgfpathlineto{\pgfqpoint{2.018182in}{1.957150in}}%
\pgfpathlineto{\pgfqpoint{2.194318in}{2.105885in}}%
\pgfpathlineto{\pgfqpoint{2.370455in}{2.075685in}}%
\pgfpathlineto{\pgfqpoint{2.546591in}{2.164775in}}%
\pgfpathlineto{\pgfqpoint{2.722727in}{2.124760in}}%
\pgfpathlineto{\pgfqpoint{2.898864in}{2.180630in}}%
\pgfpathlineto{\pgfqpoint{3.075000in}{2.176100in}}%
\pgfpathlineto{\pgfqpoint{3.251136in}{2.210075in}}%
\pgfpathlineto{\pgfqpoint{3.427273in}{2.182140in}}%
\pgfpathlineto{\pgfqpoint{3.603409in}{2.221400in}}%
\pgfpathlineto{\pgfqpoint{3.779545in}{2.213095in}}%
\pgfpathlineto{\pgfqpoint{3.955682in}{2.180630in}}%
\pgfpathlineto{\pgfqpoint{4.131818in}{2.237255in}}%
\pgfpathlineto{\pgfqpoint{4.307955in}{2.230460in}}%
\pgfpathlineto{\pgfqpoint{4.484091in}{2.242540in}}%
\pgfpathlineto{\pgfqpoint{4.660227in}{2.155715in}}%
\pgfpathlineto{\pgfqpoint{4.836364in}{2.233480in}}%
\pgfpathlineto{\pgfqpoint{5.012500in}{2.258395in}}%
\pgfpathlineto{\pgfqpoint{5.188636in}{2.256885in}}%
\pgfusepath{stroke}%
\end{pgfscope}%
\begin{pgfscope}%
\pgfpathrectangle{\pgfqpoint{0.750000in}{0.500000in}}{\pgfqpoint{4.650000in}{3.020000in}}%
\pgfusepath{clip}%
\pgfsetrectcap%
\pgfsetroundjoin%
\pgfsetlinewidth{1.505625pt}%
\definecolor{currentstroke}{rgb}{1.000000,0.498039,0.054902}%
\pgfsetstrokecolor{currentstroke}%
\pgfsetdash{}{0pt}%
\pgfpathmoveto{\pgfqpoint{0.961364in}{1.869570in}}%
\pgfpathlineto{\pgfqpoint{1.137500in}{1.757830in}}%
\pgfpathlineto{\pgfqpoint{1.313636in}{1.854470in}}%
\pgfpathlineto{\pgfqpoint{1.489773in}{1.993390in}}%
\pgfpathlineto{\pgfqpoint{1.665909in}{1.825025in}}%
\pgfpathlineto{\pgfqpoint{1.842045in}{1.908075in}}%
\pgfpathlineto{\pgfqpoint{2.018182in}{1.984330in}}%
\pgfpathlineto{\pgfqpoint{2.194318in}{1.969230in}}%
\pgfpathlineto{\pgfqpoint{2.370455in}{1.978290in}}%
\pgfpathlineto{\pgfqpoint{2.546591in}{2.133065in}}%
\pgfpathlineto{\pgfqpoint{2.722727in}{2.080970in}}%
\pgfpathlineto{\pgfqpoint{2.898864in}{2.139105in}}%
\pgfpathlineto{\pgfqpoint{3.075000in}{2.135330in}}%
\pgfpathlineto{\pgfqpoint{3.251136in}{2.170060in}}%
\pgfpathlineto{\pgfqpoint{3.427273in}{2.091540in}}%
\pgfpathlineto{\pgfqpoint{3.603409in}{2.120985in}}%
\pgfpathlineto{\pgfqpoint{3.779545in}{2.234235in}}%
\pgfpathlineto{\pgfqpoint{3.955682in}{2.200260in}}%
\pgfpathlineto{\pgfqpoint{4.131818in}{2.228950in}}%
\pgfpathlineto{\pgfqpoint{4.307955in}{2.219890in}}%
\pgfpathlineto{\pgfqpoint{4.484091in}{2.151940in}}%
\pgfpathlineto{\pgfqpoint{4.660227in}{2.275760in}}%
\pgfpathlineto{\pgfqpoint{4.836364in}{2.182895in}}%
\pgfpathlineto{\pgfqpoint{5.012500in}{2.187425in}}%
\pgfpathlineto{\pgfqpoint{5.188636in}{2.246315in}}%
\pgfusepath{stroke}%
\end{pgfscope}%
\begin{pgfscope}%
\pgfpathrectangle{\pgfqpoint{0.750000in}{0.500000in}}{\pgfqpoint{4.650000in}{3.020000in}}%
\pgfusepath{clip}%
\pgfsetrectcap%
\pgfsetroundjoin%
\pgfsetlinewidth{1.505625pt}%
\definecolor{currentstroke}{rgb}{0.172549,0.627451,0.172549}%
\pgfsetstrokecolor{currentstroke}%
\pgfsetdash{}{0pt}%
\pgfpathmoveto{\pgfqpoint{0.961364in}{1.834085in}}%
\pgfpathlineto{\pgfqpoint{1.137500in}{1.817475in}}%
\pgfpathlineto{\pgfqpoint{1.313636in}{1.904300in}}%
\pgfpathlineto{\pgfqpoint{1.489773in}{1.916380in}}%
\pgfpathlineto{\pgfqpoint{1.665909in}{1.981310in}}%
\pgfpathlineto{\pgfqpoint{1.842045in}{2.000940in}}%
\pgfpathlineto{\pgfqpoint{2.018182in}{2.046995in}}%
\pgfpathlineto{\pgfqpoint{2.194318in}{2.019815in}}%
\pgfpathlineto{\pgfqpoint{2.370455in}{2.134575in}}%
\pgfpathlineto{\pgfqpoint{2.546591in}{2.040200in}}%
\pgfpathlineto{\pgfqpoint{2.722727in}{2.108905in}}%
\pgfpathlineto{\pgfqpoint{2.898864in}{2.191955in}}%
\pgfpathlineto{\pgfqpoint{3.075000in}{2.151185in}}%
\pgfpathlineto{\pgfqpoint{3.251136in}{2.080970in}}%
\pgfpathlineto{\pgfqpoint{3.427273in}{2.222155in}}%
\pgfpathlineto{\pgfqpoint{3.603409in}{2.183650in}}%
\pgfpathlineto{\pgfqpoint{3.779545in}{2.169305in}}%
\pgfpathlineto{\pgfqpoint{3.955682in}{2.213095in}}%
\pgfpathlineto{\pgfqpoint{4.131818in}{2.210075in}}%
\pgfpathlineto{\pgfqpoint{4.307955in}{2.238765in}}%
\pgfpathlineto{\pgfqpoint{4.484091in}{2.281800in}}%
\pgfpathlineto{\pgfqpoint{4.660227in}{2.249335in}}%
\pgfpathlineto{\pgfqpoint{4.836364in}{2.262925in}}%
\pgfpathlineto{\pgfqpoint{5.012500in}{2.246315in}}%
\pgfpathlineto{\pgfqpoint{5.188636in}{2.281800in}}%
\pgfusepath{stroke}%
\end{pgfscope}%
\begin{pgfscope}%
\pgfpathrectangle{\pgfqpoint{0.750000in}{0.500000in}}{\pgfqpoint{4.650000in}{3.020000in}}%
\pgfusepath{clip}%
\pgfsetbuttcap%
\pgfsetroundjoin%
\pgfsetlinewidth{1.505625pt}%
\definecolor{currentstroke}{rgb}{0.000000,0.000000,1.000000}%
\pgfsetstrokecolor{currentstroke}%
\pgfsetdash{{5.550000pt}{2.400000pt}}{0.000000pt}%
\pgfpathmoveto{\pgfqpoint{0.750000in}{1.786736in}}%
\pgfpathlineto{\pgfqpoint{5.413889in}{1.786736in}}%
\pgfusepath{stroke}%
\end{pgfscope}%
\begin{pgfscope}%
\pgfpathrectangle{\pgfqpoint{0.750000in}{0.500000in}}{\pgfqpoint{4.650000in}{3.020000in}}%
\pgfusepath{clip}%
\pgfsetbuttcap%
\pgfsetroundjoin%
\pgfsetlinewidth{1.505625pt}%
\definecolor{currentstroke}{rgb}{0.000000,0.500000,0.000000}%
\pgfsetstrokecolor{currentstroke}%
\pgfsetdash{{5.550000pt}{2.400000pt}}{0.000000pt}%
\pgfpathmoveto{\pgfqpoint{0.750000in}{2.717183in}}%
\pgfpathlineto{\pgfqpoint{5.413889in}{2.717183in}}%
\pgfusepath{stroke}%
\end{pgfscope}%
\begin{pgfscope}%
\pgfsetrectcap%
\pgfsetmiterjoin%
\pgfsetlinewidth{0.803000pt}%
\definecolor{currentstroke}{rgb}{0.000000,0.000000,0.000000}%
\pgfsetstrokecolor{currentstroke}%
\pgfsetdash{}{0pt}%
\pgfpathmoveto{\pgfqpoint{0.750000in}{0.500000in}}%
\pgfpathlineto{\pgfqpoint{0.750000in}{3.520000in}}%
\pgfusepath{stroke}%
\end{pgfscope}%
\begin{pgfscope}%
\pgfsetrectcap%
\pgfsetmiterjoin%
\pgfsetlinewidth{0.803000pt}%
\definecolor{currentstroke}{rgb}{0.000000,0.000000,0.000000}%
\pgfsetstrokecolor{currentstroke}%
\pgfsetdash{}{0pt}%
\pgfpathmoveto{\pgfqpoint{5.400000in}{0.500000in}}%
\pgfpathlineto{\pgfqpoint{5.400000in}{3.520000in}}%
\pgfusepath{stroke}%
\end{pgfscope}%
\begin{pgfscope}%
\pgfsetrectcap%
\pgfsetmiterjoin%
\pgfsetlinewidth{0.803000pt}%
\definecolor{currentstroke}{rgb}{0.000000,0.000000,0.000000}%
\pgfsetstrokecolor{currentstroke}%
\pgfsetdash{}{0pt}%
\pgfpathmoveto{\pgfqpoint{0.750000in}{0.500000in}}%
\pgfpathlineto{\pgfqpoint{5.400000in}{0.500000in}}%
\pgfusepath{stroke}%
\end{pgfscope}%
\begin{pgfscope}%
\pgfsetrectcap%
\pgfsetmiterjoin%
\pgfsetlinewidth{0.803000pt}%
\definecolor{currentstroke}{rgb}{0.000000,0.000000,0.000000}%
\pgfsetstrokecolor{currentstroke}%
\pgfsetdash{}{0pt}%
\pgfpathmoveto{\pgfqpoint{0.750000in}{3.520000in}}%
\pgfpathlineto{\pgfqpoint{5.400000in}{3.520000in}}%
\pgfusepath{stroke}%
\end{pgfscope}%
\begin{pgfscope}%
\definecolor{textcolor}{rgb}{0.000000,0.000000,0.000000}%
\pgfsetstrokecolor{textcolor}%
\pgfsetfillcolor{textcolor}%
\pgftext[x=3.075000in,y=3.603333in,,base]{\color{textcolor}\sffamily\fontsize{12.000000}{14.400000}\selectfont CIFAR10  VGG16 (Cross-entropy loss) }%
\end{pgfscope}%
\begin{pgfscope}%
\pgfsetbuttcap%
\pgfsetmiterjoin%
\definecolor{currentfill}{rgb}{1.000000,1.000000,1.000000}%
\pgfsetfillcolor{currentfill}%
\pgfsetfillopacity{0.800000}%
\pgfsetlinewidth{1.003750pt}%
\definecolor{currentstroke}{rgb}{0.800000,0.800000,0.800000}%
\pgfsetstrokecolor{currentstroke}%
\pgfsetstrokeopacity{0.800000}%
\pgfsetdash{}{0pt}%
\pgfpathmoveto{\pgfqpoint{0.847222in}{0.569444in}}%
\pgfpathlineto{\pgfqpoint{2.858822in}{0.569444in}}%
\pgfpathquadraticcurveto{\pgfqpoint{2.886600in}{0.569444in}}{\pgfqpoint{2.886600in}{0.597222in}}%
\pgfpathlineto{\pgfqpoint{2.886600in}{1.532407in}}%
\pgfpathquadraticcurveto{\pgfqpoint{2.886600in}{1.560185in}}{\pgfqpoint{2.858822in}{1.560185in}}%
\pgfpathlineto{\pgfqpoint{0.847222in}{1.560185in}}%
\pgfpathquadraticcurveto{\pgfqpoint{0.819444in}{1.560185in}}{\pgfqpoint{0.819444in}{1.532407in}}%
\pgfpathlineto{\pgfqpoint{0.819444in}{0.597222in}}%
\pgfpathquadraticcurveto{\pgfqpoint{0.819444in}{0.569444in}}{\pgfqpoint{0.847222in}{0.569444in}}%
\pgfpathclose%
\pgfusepath{stroke,fill}%
\end{pgfscope}%
\begin{pgfscope}%
\pgfsetrectcap%
\pgfsetroundjoin%
\pgfsetlinewidth{1.505625pt}%
\definecolor{currentstroke}{rgb}{0.121569,0.466667,0.705882}%
\pgfsetstrokecolor{currentstroke}%
\pgfsetdash{}{0pt}%
\pgfpathmoveto{\pgfqpoint{0.875000in}{1.456018in}}%
\pgfpathlineto{\pgfqpoint{1.152778in}{1.456018in}}%
\pgfusepath{stroke}%
\end{pgfscope}%
\begin{pgfscope}%
\definecolor{textcolor}{rgb}{0.000000,0.000000,0.000000}%
\pgfsetstrokecolor{textcolor}%
\pgfsetfillcolor{textcolor}%
\pgftext[x=1.263889in,y=1.407407in,left,base]{\color{textcolor}\sffamily\fontsize{10.000000}{12.000000}\selectfont Self learning run\# 1}%
\end{pgfscope}%
\begin{pgfscope}%
\pgfsetrectcap%
\pgfsetroundjoin%
\pgfsetlinewidth{1.505625pt}%
\definecolor{currentstroke}{rgb}{1.000000,0.498039,0.054902}%
\pgfsetstrokecolor{currentstroke}%
\pgfsetdash{}{0pt}%
\pgfpathmoveto{\pgfqpoint{0.875000in}{1.266203in}}%
\pgfpathlineto{\pgfqpoint{1.152778in}{1.266203in}}%
\pgfusepath{stroke}%
\end{pgfscope}%
\begin{pgfscope}%
\definecolor{textcolor}{rgb}{0.000000,0.000000,0.000000}%
\pgfsetstrokecolor{textcolor}%
\pgfsetfillcolor{textcolor}%
\pgftext[x=1.263889in,y=1.217592in,left,base]{\color{textcolor}\sffamily\fontsize{10.000000}{12.000000}\selectfont Self learning run\# 2}%
\end{pgfscope}%
\begin{pgfscope}%
\pgfsetrectcap%
\pgfsetroundjoin%
\pgfsetlinewidth{1.505625pt}%
\definecolor{currentstroke}{rgb}{0.172549,0.627451,0.172549}%
\pgfsetstrokecolor{currentstroke}%
\pgfsetdash{}{0pt}%
\pgfpathmoveto{\pgfqpoint{0.875000in}{1.076389in}}%
\pgfpathlineto{\pgfqpoint{1.152778in}{1.076389in}}%
\pgfusepath{stroke}%
\end{pgfscope}%
\begin{pgfscope}%
\definecolor{textcolor}{rgb}{0.000000,0.000000,0.000000}%
\pgfsetstrokecolor{textcolor}%
\pgfsetfillcolor{textcolor}%
\pgftext[x=1.263889in,y=1.027778in,left,base]{\color{textcolor}\sffamily\fontsize{10.000000}{12.000000}\selectfont Self learning run\# 3}%
\end{pgfscope}%
\begin{pgfscope}%
\pgfsetbuttcap%
\pgfsetroundjoin%
\pgfsetlinewidth{1.505625pt}%
\definecolor{currentstroke}{rgb}{0.000000,0.000000,1.000000}%
\pgfsetstrokecolor{currentstroke}%
\pgfsetdash{{5.550000pt}{2.400000pt}}{0.000000pt}%
\pgfpathmoveto{\pgfqpoint{0.875000in}{0.886574in}}%
\pgfpathlineto{\pgfqpoint{1.152778in}{0.886574in}}%
\pgfusepath{stroke}%
\end{pgfscope}%
\begin{pgfscope}%
\definecolor{textcolor}{rgb}{0.000000,0.000000,0.000000}%
\pgfsetstrokecolor{textcolor}%
\pgfsetfillcolor{textcolor}%
\pgftext[x=1.263889in,y=0.837963in,left,base]{\color{textcolor}\sffamily\fontsize{10.000000}{12.000000}\selectfont 4000-label 77.04 \(\displaystyle \pm\) 0.97\%}%
\end{pgfscope}%
\begin{pgfscope}%
\pgfsetbuttcap%
\pgfsetroundjoin%
\pgfsetlinewidth{1.505625pt}%
\definecolor{currentstroke}{rgb}{0.000000,0.500000,0.000000}%
\pgfsetstrokecolor{currentstroke}%
\pgfsetdash{{5.550000pt}{2.400000pt}}{0.000000pt}%
\pgfpathmoveto{\pgfqpoint{0.875000in}{0.696759in}}%
\pgfpathlineto{\pgfqpoint{1.152778in}{0.696759in}}%
\pgfusepath{stroke}%
\end{pgfscope}%
\begin{pgfscope}%
\definecolor{textcolor}{rgb}{0.000000,0.000000,0.000000}%
\pgfsetstrokecolor{textcolor}%
\pgfsetfillcolor{textcolor}%
\pgftext[x=1.263889in,y=0.648148in,left,base]{\color{textcolor}\sffamily\fontsize{10.000000}{12.000000}\selectfont All label 89.37 \(\displaystyle \pm\) 0.49\%}%
\end{pgfscope}%
\end{pgfpicture}%
\makeatother%
\endgroup%